%% file: Formatting-Instructions-LaTeX-2026.tex
\documentclass[letterpaper]{article} % DO NOT CHANGE THIS
\usepackage{aaai2026}  % DO NOT CHANGE THIS
\usepackage{times}  % DO NOT CHANGE THIS
\usepackage{helvet}  % DO NOT CHANGE THIS
\usepackage{courier}  % DO NOT CHANGE THIS
\usepackage[hyphens]{url}  % DO NOT CHANGE THIS
\usepackage{graphicx} % DO NOT CHANGE THIS
\usepackage{natbib}  % DO NOT CHANGE THIS AND DO NOT ADD ANY OPTIONS TO IT
\usepackage{caption} % DO NOT CHANGE THIS AND DO NOT ADD ANY OPTIONS TO IT
\usepackage{algorithm}
\usepackage{algorithmic}
\usepackage{amsmath}
\usepackage{subcaption}
\usepackage{booktabs}
\usepackage{pifont}
\usepackage[table]{xcolor}
\usepackage{multirow}
\usepackage{amssymb}
\usepackage{listings}
\DeclareCaptionStyle{ruled}{labelfont=normalfont,labelsep=colon,strut=off} % DO NOT CHANGE THIS
\floatstyle{ruled}
\newfloat{listing}{tb}{lst}{}
\floatname{listing}{Listing}
\usepackage{soul}

\title{LAMP: Learning Universal Adversarial Perturbations for Multi-Image Tasks via Pre-trained Models}
\author{
    Alvi Md Ishmam,
    Najibul Haque Sarker,
    Zaber Ibn Abdul Hakim,
    Chris Thomas
}
\affiliations{
    Department of Computer Science, Virginia Tech\\
    alvi@vt.edu, najibulhaque@vt.edu, zaberhakim@vt.edu, christhomas@vt.edu
}

\usepackage{bibentry}
\begin{document}

\maketitle

\begin{abstract}
Multimodal Large Language Models (MLLMs) have achieved remarkable performance across vision-language tasks. Recent advancements allow these models to process multiple images as inputs. However, the vulnerabilities of multi-image MLLMs remain unexplored. Existing adversarial attacks focus on single-image settings and often assume a white-box threat model, which is impractical in many real-world scenarios. 
This paper introduces LAMP, a black-box method for learning Universal Adversarial Perturbations (UAPs) targeting multi-image MLLMs. LAMP applies an attention-based constraint that prevents the model from effectively aggregating information across images. LAMP also introduces a novel cross-image contagious constraint that forces perturbed tokens to influence clean tokens, spreading adversarial effects without requiring all inputs to be modified. 
Additionally, an index-attention suppression loss enables a robust position-invariant attack. Experimental results show that LAMP outperforms SOTA baselines and achieves the highest attack success rates across multiple vision-language tasks and models.

\end{abstract}


% \input{AnonymousSubmission/LaTeX/sections/intro}
\section{Introduction}
\label{sec:intro}

\begin{figure}[ht!]
    \centering
    \includegraphics[width=\linewidth]{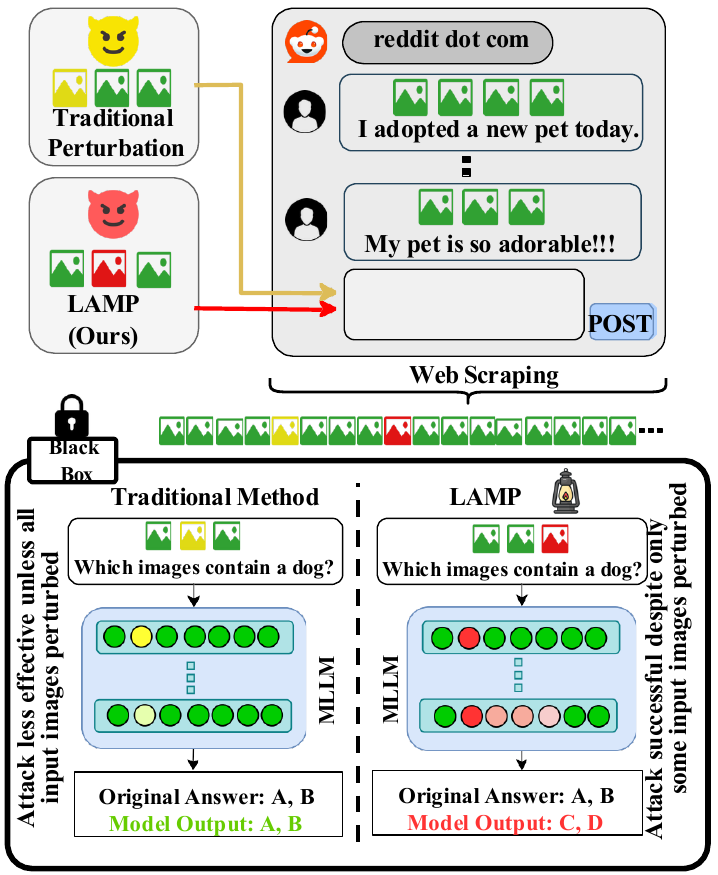}
     \caption{An overview of our approach showing superior effectiveness over traditional methods. Conventional methods fail when the perturbation is not applied to every image, an unrealistic setting when the attacker does not have access to the later inference stage. Our method succeeds even if a subset of downstream samples includes perturbed samples since it can affect ``green'' samples even though they are not attacked, unlike traditional methods.}
    \label{fig:concept}

\end{figure}

Multimodal Large Language Models (MLLMs) like GPT-4V \cite{gpt4v}, Gemini \cite{gemini}, LLaVA-NeXT \cite{llava}, and Idefics \cite{laurenccon2024matters} have made significant advancements in visual-language understanding and generation, particularly for single-image tasks such as VQA \cite{vqa}.  A few open-source models, such as Mantis \cite{mantis} and VILA \cite{vila}, extend these capabilities to multi-image inputs, enabling coreference, comparison, reasoning, and temporal understanding. These models first learn multimodal interactions through pre-training on unlabeled image-text datasets. They are then fine-tuned using labeled image-text pairs for various multi-image downstream tasks \cite{nlvr2, dreamsim, next}. Despite their remarkable performance, the adversarial robustness of multi-image MLLMs remains unexplored.

Several studies have evaluated the robustness of single-image vision-language models. Most existing approaches \cite{adv_rob, cross} focus on white-box attacks, assuming access to gradient information from fine-tuned models. In practice, however, attackers typically only have access to public pretrained models and lack knowledge of downstream models. Gradient-based white-box methods \cite{pgd} also generate instance-specific perturbations that generalize poorly and require costly re-optimization for new inputs.

Although some efforts, such as AdvCLIP \cite{advclip}, have explored this challenge, existing approaches are often model-specific or lack practical utility due to limited imperceptibility. Additionally, generating input-specific adversarial perturbations for multi-image multimodal large language models (MLLMs) using methods like \citet{pgd} is impractical. A scenario is shown in Fig.~\ref{fig:concept}. Consider an attacker who is posting images on social media but cannot control which sets of images, how many, or in what order they are fed into the model. At best, with current methods, an attacker can post multiple images with an attack using existing techniques. However, these attacks are not learned jointly to have a synergistic effect.
 
In this paper, we address that limitation by developing a new multi-image attack that works synergistically under real-world conditions, where the attacker cannot control how many attack images are presented to the model or in what order. Unlike multiple single image attacks, our attack is explicitly designed to attack multi-image scenarios. Traditional methods, designed primarily for single-image attacks, are thus less effective in multi-image contexts, unless all images in the instance are perturbed. To address this, we propose a perturbation learning framework capable of carrying out effective attacks even when a small subset of perturbed images is present within the inference instance. Our approach is designed to propagate the perturbation effect across subsequent tokens, allowing the adversarial influence to persist throughout the model’s generation process, which enables a successful attack under more realistic constraints.

To address these challenges, we propose a novel method for generating Universal Adversarial Perturbations (UAPs) that targets black-box models. Our approach learns UAPs using a pretrained model and attacks various multi-image MLLMs without prior knowledge of their architectures or downstream tasks (Fig.~\ref{fig:concept}). In many real-world scenarios (e.g.~an attacker serving poisoned ad images displayed on a webpage processed by a model), attackers cannot poison all images or control how many images the model ingests during inference. Existing methods assume a single perturbed image, which does not exploit the unique attack surface posed in the multi-image setting. Our approach learns UAPs specifically for these settings by maximizing the dissimilarity between clean and perturbed images. It disrupts attention weights between them while keeping the pretrained MLLMs frozen. This ensures that the learned UAPs remain effective and transferable across different tasks and models. To do so, first we train UAPs by minimizing the probability of correct predictions. Next, we enforce dissimilarity constraints between the hidden states of each decoder layer in the LLM backbone. We also constrain the dissimilarity of attention weights using the Pompeiu-Hausdorff distance \cite{phd_distance} to learn attacks which target specific heads.
A fixed number of UAPs may not generalize well to varying numbers of images during inference. To address this, we introduce a novel ``contagious'' objective that encourages a fixed set of perturbed tokens to focus more on clean tokens in the self-attention space. This technique allows an attacker to induce noisy behavior in clean images without knowing how many perturbations each sample requires. Moreover, we propose an index-attention suppression constraint to enable position-invariant attacks tailored to multi-image settings. The contributions of our works:
\begin{itemize}
    \item We propose the first adversarial attack targeting multi-image MLLMs to our knowledge, exploiting the unique attack surface enabled by multiple inputs. Our attack is transferable across MLLMs without requiring UAPs tailored to specific downstream models or  tasks.
    \item We introduce a novel method for learning UAPs by leveraging the LLM's self-attention module without optimizing the MLLM itself by using Pompeiu-Hausdorff distance \cite{phd_distance}. 
    \item We propose a novel ``contagious" objective that enables perturbed visual tokens to infect clean tokens, allowing for the learning of a fixed number of UAPs in multi-image settings. We also propose an ``index-attention suppression" loss, that enables the position-invariant attacks.   
    \item We conduct a comprehensive experimental evaluation across a wide range of MLLMs, challenging multi-image benchmark datasets along with VQA and image captioning tasks. Our results demonstrate that our attack method significantly outperforms SOTA approaches.
\end{itemize}

\section{Related Work}
\label{sec:related}

% \textbf{Single-Modal Adversarial Attacks.} 
% Prior adversarial attack methods leveraged gradient-based perturbations \cite{goodfellow2014explaining, madry2017towards}, initially targeting image classification models \cite{moosavi2016deepfool, nguyen2015deep}, and later extended to object detection and segmentation \cite{xie2017adversarial}. While first explored in vision, adversarial attacks were also adapted to text, with perturbations applied at the character \cite{ebrahimi2017hotflip} and embedding levels \cite{miyato2016adversarial}.
 % Prior adversarial attack methods explored using the sign of the gradients and its variants to generate adversarial perturbations \cite{goodfellow2014explaining, madry2017towards}. Initial research predominantly targeted image classification models \cite{moosavi2016deepfool, nguyen2015deep}; however, subsequent studies demonstrated the vulnerability of object detection and segmentation models to adversarial attacks \cite{xie2017adversarial}. While these attacks were first introduced in the visual domain, they were later extended to text-based models as well. In the textual domain, adversarial perturbations have been applied at various levels, including character \cite{ebrahimi2017hotflip}, word \cite{liang2017deep}, and word embedding perturbations \cite{miyato2016adversarial}.

\noindent\textbf{Multi-Modal Adversarial Attacks.} 
With the increasing popularity of vision-language models such as CLIP \cite{radford2021learning}, BLIP \cite{li2022blip}, 
researchers have focused on assessing their robustness by developing adversarial attack strategies. \citet{xu2018fooling} demonstrated that iterative pixel-level perturbations can effectively deceive visual question answering models. Expanding on this, \citet{agrawal2018don, shah2019cycle} introduced attacks targeting the textual modality of multimodal models. More recent approaches, such as Co-Attack \cite{zhang2022towards} and SGA \cite{set-level}, explore joint perturbations across visual and textual modalities.

% \noindent\textbf{Conventional Adversarial Attacks.} 

\noindent\textbf{Universal Adversarial Attacks.}
Adversarial attack research has primarily focused on instance-specific methods in both single-modal \cite{szegedy2013intriguing, kim2019single} and multi-modal \cite{xu2018fooling, zhang2022towards} settings under white and black-box assumptions. In contrast, universal adversarial perturbations (UAPs) \cite{moosavi2017universal, mopuri2018generalizable} offer a more practical, sample-agnostic alternative. While UAPs have been widely studied in image \cite{hayes2018learning, khrulkov2018art} and text \cite{xue2024trojllm, wallace2019universal} domains, their application to vision-language pretrained (VLP) and multimodal large language models (MLLMs) remains limited. UAP-VLP \cite{uap_vlp} and Doubly-UAP \cite{Doubly-UAP} target image-only UAPs through sub-region optimization and attention manipulation. CPGC-UAP \cite{cpgc} extends UAPs to both modalities using a generator, while DO-UAP \cite{yang2024efficient} employs direct optimization for efficiency but is limited to single-image inputs. Jailbreak-MLLM \cite{jailbreak-mllm} improves transferability by attacking MLLM ensembles.

\noindent\textbf{Multi-Image Adversarial Attacks.} Existing methods take advantage of MLLM multi-image capability for composite adversarial attacks. AnyDoor \cite{lu2024test} shows the effectiveness of UAPs when attacking randomly selected frames in a video. Multiple scenario-aware adversarial images are generated and used as a collaborative adversarial attack in MLAI \cite{hao2025exploring}. On the other hand, \citet{broomfielddecompose} splits harmful texts into multiple typographic images to leverage multi-image capabilities of MLLMs. \citet{wang2025align} explores multi-modal in-context attacks by providing few-shot adversarial images and texts as context. However, none of these methods consider universal adversarial attack methods on subsets of interleaved images in MLLMs.

\section{Problem Setting}
\label{sec:problem}

\begin{figure*}[t]
    \centering
    \includegraphics[width=1\linewidth]{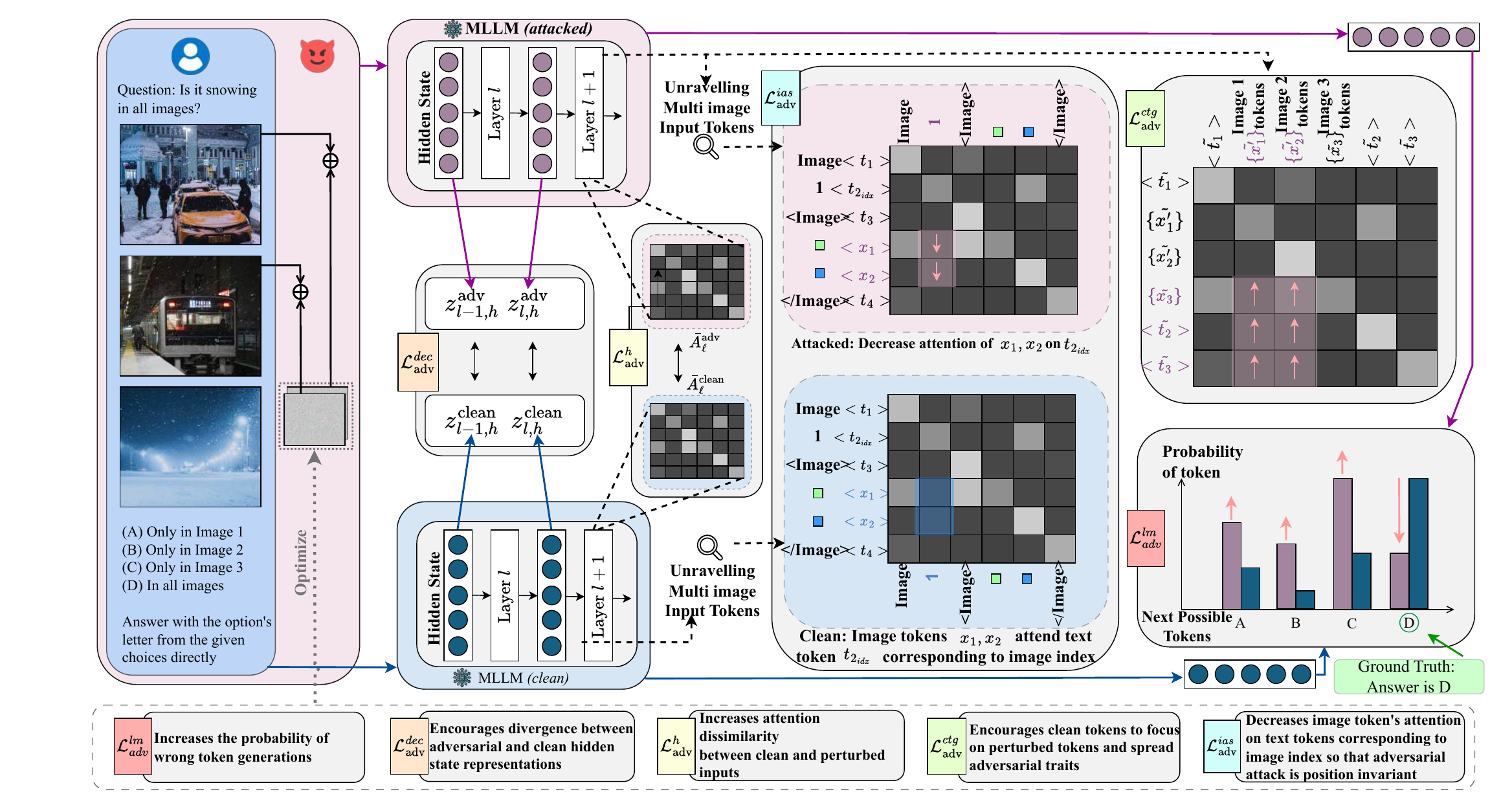}
    \caption{An overview of our proposed attack methodology. The input sample shown in the left \textcolor[HTML]{81ACD7}{blue}-shaded box is what a normal user might query an MLLM, while the \textcolor[HTML]{D7ACC1}{pink} box shows our attack setting, where the attacker adds learned universal perturbations to two images of the input. The universal perturbations are learned using: \textbf{a) Adversarial language modeling loss $\mathcal{L}_{adv}^{lm}$:} reduces likelihood of correct tokens (Option: D), and increases probability of wrong tokens (Options: A, B, C). \textbf{b) Adversarial hidden states loss $\mathcal{L}_{adv}^{dec}$:} encourages divergence between \( z_{l,h}^{\text{adv}} \) and \( z_{l,h}^{\text{clean}} \), representing the hidden states of the \( h \)-th attention head in the \( l \)-th decoder layer for adversarial and clean inputs. \textbf{c) Adversarial attention weights loss $\mathcal{L}_{adv}^{h}$:} maximizes distance between $\bar{A}^{\text{clean}}_{\ell}$ and $\bar{A}^{\text{adv}}_{\ell}$, representing head-averaged attention weights in \( l \)-th decoder layer for adversarial and clean inputs. \textbf{d) Adversarial contagious loss $\mathcal{L}_{adv}^{ctg}$:} encourages clean tokens to place greater attention to noisy image tokens $\tilde{x'_1}$ and $\tilde{x'_2}$ for each $A^{(l)}_{:,h}$, attention weights for head $h$ at layer $l$ (Here $\{x_t\}$ represents all image tokens of image $x_t$ for brevity). And \textbf{e) Adversarial Index-Attention Suppression loss $\mathcal{L}_{adv}^{ias}$:} suppresses attention from image tokens $x_1, x_2$ to text tokens corresponding to image index $t_{2_{idx}}$, to encourage image position invariance (Here the input token sequence for multi-image setting is shown as `Image 1: \textless Image\textgreater \textless image\textgreater\textless/Image\textgreater').}
    \label{fig:method2}
\end{figure*}

\subsection{Threat Model}

\textbf{Adversary Objective.} Given a pretrained MLLM $\mathcal{M}$, our goal is to learn imperceptible UAPs that can transfer across different downstream MLLMs and tasks. In this setting, the target MLLMs, datasets, and downstream tasks remain unknown during training, and the attacker cannot control learned text embeddings. For instance, a malicious actor could serve image-based advertisements containing adversarial noise or post adversarially perturbed images in online comments. When a model processes a webpage containing multiple images, the attack could still be effective even if the adversary does not control all images or associated text. To handle this black-box scenario, we learn adversarial perturbations using a surrogate dataset and model $\mathcal{M}$. The surrogate multimodal dataset is denoted as $\mathcal{D}_s = \{(x^{(i)}, t^{(i)})\}_{i=1}^{n}$, where $\mathcal{D}_s$ consists of $n$ multimodal samples. Here, $x^{(i)} = \{x_j^{(i)}\}_{j=1}^{m^{(i)}}$ represents the set of $m^{(i)}$ images for the $i$-th sample, and $t^{(i)}$ is the corresponding text prompt. The objective is to learn imperceptible universal adversarial perturbations $\delta_1, \dots, \delta_k$ where $\|\delta_k\|_{\infty} \leq \epsilon$; $\epsilon$ is the perturbation budget and $l_{\infty}$ denotes the perturbation constraint. Here, we attack subset of images in sample $m^{(i)}$, where $k < m^{(i)}$. We consider a multi-image setting where a user is expected to provide multiple images when querying a target MLLM. An attacker can use the learned UAPs to corrupt a subset of the user’s images, misleading multimodal models into making incorrect image–prompt associations and producing erroneous responses at inference.

\textbf{Adversary Capabilities.}  In a black-box setting, an adversary has no control over fine-tuned models and does not know how many images are used per sample during inference. Traditional perturbation generation methods require direct access to fine-tuned models and datasets, allowing adversaries to perturb specific target images -- an impractical scenario for MLLMs trained on web-scale data. Our setting is more realistic, as we aim to learn only a small, fixed number of universal perturbations that effectively attack multi-image MLLMs. We assume a black-box scenario where the adversary has no knowledge of the target model’s architecture or training process. Most importantly, since the number of images used during inference is unknown, our approach ensures that perturbing a fixed number of images can still generate a strong and transferable attack.

\section{Attack Methodology}
\label{sec:method}
Our method proposes to learn UAPs using an accessible pretrained model, which can then be applied to black-box target models. We impose a constraint on the language model head by reducing the probability of the correct token. Additionally, we introduce a constraint to increase the divergence between the hidden states of perturbed and clean inputs in the LLM decoder. We also add a Pompeiu-Hausdorff distance \cite{phd_distance} based constraint between the clean and perturbed attention weights. Furthermore, we encourage the model to allocate more attention to perturbed tokens from clean tokens through a novel "contagious" objective and an index-attention suppression objective (Fig. \ref{fig:method2}).

\subsection{Adversarial language modeling loss}
We apply adversarial perturbations to a subset of images in an input sequence of an MLLM. Let a sample contain $M$ images and a corresponding text prompt $t$. The images are $x_1, x_2, \dots, x_M$. We introduce learnable adversarial perturbations $\delta_1, \delta_2, \dots, \delta_k$, constrained by \( \|\delta_k\|_{\infty} \leq \epsilon \), to generate perturbed images. The perturbed image, 
${x'_k} = x_k + \delta_k$,  $k<M$. The final interleaved input sequence is:
$s = (\tilde{x'_1}, \tilde{x'_2}, \tilde{x_3}, \dots, \tilde{x_m}, \tilde t_1, \tilde t_2, \dots, \tilde t_n)$. Here, $\tilde{x'_1}$,  $\tilde{x'_2}$ represents image tokens' of adversarial images ${x'_1}$ and ${x'_2}$ respectively. $\tilde{x_3} \dots \tilde{x_m}$ are clean image tokens. For brevity, we skipped all tokens per image. The adversarial language modeling loss is:
\begin{equation}
\label{eqn:adv_lm}
        \mathcal{L}_{adv}^{lm} = - \frac{1}{N} \sum_{i=1}^{N} \log (1 - P_{\theta}(t_{i+1} | s_{1:i})),
\end{equation}

where \( P_{\theta}(t_{i+1} | s_{1:i}) \) is the predicted probability of the correct token and $N$ number of tokens in the sequence. Through the loss, we encourage the model to reduce the likelihood of correct tokens, increasing the probability of wrong token generations while optimizing  {$\delta_k$}
keeping the MLLM parameters frozen. Note that the summation over batch is omitted for brevity.

\subsection{Adversarial hidden states loss}

We introduce a loss function to learn the adversarial perturbations that maximize the cosine distance between the hidden states across decoder layers and attention heads. Let \( z_{l}^{\text{adv}} \) and \( z_{l}^{\text{clean}} \) represent mean hidden state in layer l, averaged over heads for adversarial and clean inputs, respectively. To encourage divergence between the adversarial and clean representations, our objective is:

% \vspace{-1em}
\begin{equation}
\label{eqn:adv_dec}
\mathcal{L}_{\text{adv}}^{dec} = \frac{1}{L} \sum_{l=1}^{L}  \cos(z_{l}^{\text{adv}}, z_{l}^{\text{clean}})
\end{equation}

where \( L\) is the total number of decoder layers, and \( H\) is the total number of attention heads per layer. By minimizing \( \mathcal{L}_{\text{adv}}^{dec}\), we push the adversarial hidden states away from their clean counterparts across all layers and layer heads. The similarity is measured by the cosine similarity $cos(\cdot,\cdot)$.

\subsection{Adversarial attention via relaxed Pompeiu-Hausdorff distance}
Attention weights indicate how tokens contribute to a model’s internal representations. Adversarial perturbations modify these patterns; hence, we amplify these changes to force distinct behaviors between clean and perturbed inputs.

The Pompeiu-Hausdorff distance \cite{phd_distance} offers a worst-case measure by quantifying the maximum deviation between two sets, i.e.~the distributions of attention weights. It is defined as:
\begingroup
\fontsize{6.5}{8}\selectfont
\begin{equation}
\begin{aligned}
\label{eqn:phd}
d_{hd}(S_1, S_2) = \max \Big\{ &kth\sup_{s_1 \in S_1} \inf_{s_2 \in S_2} d(s_1, s_2), 
kth \sup_{s_2 \in S_2} \inf_{s_1 \in S_1} d(s_2, s_1) \Big\}
\end{aligned}
\end{equation}
\endgroup
where \( d(s_1, s_2) \) is a distance metric (e.g. Euclidean or cosine) and $\sup\limits_{s_1 \in S_1} \inf\limits_{s_2 \in S_2}$ means selecting $kth$ maximum value in set $D_1 = {\min\limits_{s_2 \in S_2} d(s_{1(i)}, s_2), s_{1(i)} \in S_1}$ and vice versa. This relaxed formulation captures worst-case local discrepancies that may be missed by measures such as KL divergence. In practical terms, we force the model to exhibit pronounced differences in its internal focus even in regions where the global distribution might otherwise appear similar. Note this differs from a simple average sum which distributes differences across all tokens. 

Let \( A^{\text{clean}}_{\ell,h} \) and \( A^{\text{adv}}_{\ell,h} \) denote the attention matrices at layer \( \ell \) and head \( h \) for clean and adversarial inputs. We first average over heads:
% \vspace{-0.5em}
\begin{equation}
\bar{A}^{\text{clean}}_{\ell} = \frac{1}{H} \sum_{h=1}^{H} A^{\text{clean}}_{\ell,h}, \quad
\bar{A}^{\text{adv}}_{\ell} = \frac{1}{H} \sum_{h=1}^{H} A^{\text{adv}}_{\ell,h}
\end{equation}
The Hausdorff distance between the averaged weights:\\[-0.75em]
% \vspace{-20pt}

\begingroup
\fontsize{6.5}{8}\selectfont
{\footnotesize
\begin{equation}
\label{eq:phd_distance}
\begin{aligned}
d_{hd}(\bar{A}^{\text{clean}}_{\ell}, \bar{A}^{\text{adv}}_{\ell}) = \max \Big\{ 
\sup_{a_c \in \bar{A}^{\text{clean}}_{\ell}} \inf_{a_a \in \bar{A}^{\text{adv}}_{\ell}} d(a_c, a_a), \\
\sup_{a_a \in \bar{A}^{\text{adv}}_{\ell}} \inf_{a_c \in \bar{A}^{\text{clean}}_{\ell}} d(a_a, a_c)
\Big\}
\end{aligned}
\end{equation}
}
\endgroup

We define the adversarial loss by averaging over all layers:
% \vspace{-0.5em}
\begin{equation}
\label{eq:phd_loss}
\mathcal{L}_{\text{adv}}^{h} = -\frac{1}{L} \sum_{\ell=1}^{L} d_{hd}(\bar{A}^{\text{clean}}_{\ell}, \bar{A}^{\text{adv}}_{\ell})
\end{equation}
Minimizing \(\mathcal{L}_{\text{adv}}^{h}\) encourages the model to exhibit significantly different internal focus when processing clean versus adversarial inputs.

\subsection{Adversarial contagious loss}
Adversarial perturbations are typically constrained to specific image inputs, but their effect can propagate across the model’s internal representations. We introduce a novel concept of \textbf{contagious} loss, leveraging the idea that adversarial perturbations can influence clean tokens through self attention mechanism. Specifically, in an adversarial sample, where some images are perturbed and some remain clean, we encourage the clean tokens to pay more attention to perturbed tokens to adopt adversarial characteristics. This idea helps us to learn a fixed number of adversarial perturbations without explicitly perturbing all images. For instance, in realistic scenarios at inference time an attacker has no idea of how many images are fed and how many perturbations are required. Let \( L \) be the number of layers in the model, \( H \) be the number of attention heads in each layer, \( \mathcal{N} \) represent the indices of noisy tokens (image tokens), and \( \mathcal{C} \) represent the indices of clean tokens. \( A^{(l)}_{:, h, i, j} \) represents the attention weight at layer \( l \), head \( h \), which shows how much clean token \( i \) contributes to the perturbed image token \( j \). We introduce loss to maximize attention weights to encourage clean image and text tokens to pay higher weights to noisy image tokens and indicate where the model should ``attend" to.

% \vspace{-2em}
\begin{equation}
\label{eq:ctg}
\mathcal{L}_{\text{adv}}^{\text{ctg}} = - \frac{1}{LH} \sum_{l=1}^{L} \sum_{h=1}^{H} \sum_{i \in \mathcal{C}} \sum_{j \in \mathcal{N}} A^{(l)}_{:, h, i, j}
\end{equation}

% The final objective is the combination of Eq. \ref{eqn:adv_lm},  \ref{eqn:adv_dec}, \ref{eq:phd_loss}, \ref{eq:ctg} and $\lambda_1, \lambda_2, \lambda_3, \lambda_4 > 0$. We refer it as LAMP.
% \begin{equation}
%     \label{eq:final_loss}
%  \arg\max_{\delta_k} \mathcal{L}_{\text{adv}} = \lambda_1\mathcal{L}_{\text{adv}}^{\text{lm}} +  \lambda_2\mathcal{L}_{\text{adv}}^{\text{dec}} + \lambda_3\mathcal{L}_{\text{adv}}^{\text{h}} + \lambda_4\mathcal{L}_{\text{adv}}^{\text{ctg}}
% \end{equation}

\subsection{Adversarial Index-Attention Suppression Loss }
%%%%%%%%%%%%%%%%
In single-image multimodal adversarial attacks, the input contains only one image placed at a fixed token position. Since its location is static, position-dependent reasoning does not influence the attack's success; the perturbation only needs to disrupt image-text alignment.

In contrast, multi-image settings involve interleaved sequences of text and images, often containing index-referencing phrases like \texttt{``image 1:"}, \texttt{``image 2:"}, etc. These index tokens provide explicit positional grounding, enabling the model to associate specific visual tokens with their corresponding references. For example, in a prompt like: ``In (image 1: \texttt{<Image><image></Image>}) and (image 2: \texttt{<Image><image></Image>}), which image shows a more economically advanced place?'', correct reasoning requires resolving visual content based on index markers.

In such settings, an adversarial image may only succeed when placed in a specific slot (e.g., the first image), because the model learns to associate index tokens (e.g., ``1'') with nearby image tokens via causal attention. Specifically, when tokens are ordered as $t_1, t_2, \dots, t_{\text{idx}}, x_1, \dots, x_m, \dots, t_n$, image tokens $x_1 \dots x_m$ may attend to their corresponding index token $t_{\text{idx}}$. This creates a positional vulnerability.

To make attacks robust to image reordering, we propose a \textit{position-invariant adversarial attack} that penalizes attention from image tokens to their associated index tokens during perturbation learning. By decoupling image tokens from their position-specific textual anchors, the attack generalizes across image positions.

Let $A^{(l)} \in \mathbb{R}^{ H \times T \times T}$ be the attention weights at layer $l \in \{1, \dots, L\}$, with $H$ heads and $T$ tokens. Let $\mathcal{I}^{(k)} \subset \{0, \dots, T-1\}$ denote the image token indices for image $k$, and $t^{(k)}_{\text{idx}}$ the corresponding index token position. The \textbf{Index-Attention Suppression Loss} is defined as:

% \vspace{-20pt}

%%%%%%%%%%%%%%%%%

\begin{equation}
\mathcal{L}_{\text{adv}}^{ias} = \frac{1}{LH}
\sum_{l=1}^{L} \sum_{h=1}^{H} \sum_{k=1}^{K} \sum_{i \in \mathcal{I}^{(k)}} A^{(l)}_{h,i,t^{(k)}_{\text{idx}}}
\end{equation}
The final objective is the combination of Eq. \ref{eqn:adv_lm},  \ref{eqn:adv_dec}, \ref{eq:phd_loss}, \ref{eq:ctg} and $\lambda_1, \lambda_2, \lambda_3, \lambda_4, \lambda_5 > 0$. We refer to it as \emph{LAMP}:
\begin{equation}
    \label{eq:final_loss}
 \mathcal{L}_{\text{adv}} = \lambda_1\mathcal{L}_{\text{adv}}^{\text{lm}} +  \lambda_2\mathcal{L}_{\text{adv}}^{\text{dec}} + \lambda_3\mathcal{L}_{\text{adv}}^{\text{h}} + \lambda_4\mathcal{L}_{\text{adv}}^{\text{ctg}} + 
\lambda_5\mathcal{L}_{\text{adv}}^{\text{ias}}
\end{equation}

% \input{AnonymousSubmission/LaTeX/sections/experiments}
% \vspace{-2em}
\section{Experiments}
\label{sec:experiments}

\subsection{Experimental Setup}

\textbf{Model and Dataset.} We use the pretrained Mantis-CLIP \cite{mantis} model to learn imperceptible perturbations since the Mantis family is the best performing open source model for multi-image tasks \cite{mantis}. Notably, the parameters of the multimodal model's image encoder and language model are kept frozen. The maximum context length is set to $8192$. We use AdamW with weight decay and a cosine scheduler, starting with a learning rate of $10^{-4}$ and a decay rate of $0.2$. Training is conducted for $20$ epochs with a batch size of $128$ on $17,000$ samples from the Mantis Instruct dataset \cite{mantis}. The learned perturbation has a shape of $336 \times 336$. For all experiments, the perturbation budget $\epsilon$ is uniformly set to $12/255$. All experiments were conducted on A100 GPUs. 

% \vspace{-2.0em}
\noindent\textbf{Evaluation Benchmarks and Target Models}.
We experiment on five multi-image benchmark tasks in total-- NLVR2 \cite{nlvr2}, and Qbench \cite{qbench} and 3 held-out benchmarks: Mantis-Eval \cite{mantis}, BLINK \cite{blink}, and MVBench \cite{mvbench}. We select multi-image MLLMs as our target model for querying these learned UAPs from the pretrained model. The target models are 
Mantis-CLIP, Mantis-SIGLIP, Mantis-Idefics2 \cite{mantis}, VILA-1.5 \cite{vila}, LLaVA-v1.6 \cite{llava}, Qwen-VL-Chat \cite{qwen}, Qwen-2.5 \cite{qwen2.5}, MiniGPT4 \cite{minigpt}. We have also experimented on single image VQA tasks MM-Vet \cite{mm-vet}, LLaVA-Bench \cite{llava-bench} and multi-image QA Mantis-Eval \cite{mantis}. 
We also experimented on the selection-free VQA (OK-VQA \cite{ok-vqa}) and image captioning MSCOCO \cite{mscoco} tasks following \cite{pandoras_box}.

\noindent\textbf{Evaluation Metrics.} We utilize  Attack Success Rate (ASR) as a metric to quantify the effectiveness of the proposed attack and baselines following prior research \cite{uap_vlp, cpgc, set-level}. ASR is calculated as the percentage of adversarial examples that successfully deceive the model by generating incorrect outputs. The higher the ASR, the better the attack performance.

\begin{table}[t]
  \centering
  \small
  \begin{tabular}{lccc}
    \toprule
    \textbf{Setting} & \textbf{Avg.\ Best} & \textbf{LAMP} & \textbf{$\Delta$} \\
                     & \textbf{Baseline (\%)} & (\%) & (pp) \\
    \midrule
    \multicolumn{4}{c}{\textit{Per Target Model}}\\
    \rowcolor{gray!20}
    Mantis-CLIP        & 51.5 & 71.9 & +20.4 \\
    Mantis-SIGLIP      & 51.6 & 71.9 & +20.3 \\
    Mantis-Idefics2    & 49.2 & 72.4 & +23.2 \\
    VILA-1.5           & 56.1 & 76.2 & +20.1 \\
    LLaVA-v1.6         & 58.5 & 78.9 & +20.4 \\
    Qwen-VL-Chat       & 64.4 & 79.9 & +15.5 \\
    Qwen-2.5           & 62.5 & 79.4 & +16.9 \\
    \cmidrule{2-4}
    \textbf{Overall}   & 56.3 & 75.8 & +19.5 \\
    \midrule\midrule
    \multicolumn{4}{c}{\textit{Per Dataset}}\\
    Mantis Eval        & 59.4 & 77.7 & +18.4 \\
    NLVR2              & 39.4 & 59.7 & +20.3 \\
    BLINK              & 66.9 & 85.7 & +18.8 \\
    Q-Bench            & 52.5 & 76.0 & +23.4 \\
    MVBench            & 63.1 & 80.0 & +16.9 \\
    \bottomrule
  \end{tabular}
  \caption{Average ASR (\%) and absolute improvement
           of \textbf{LAMP} over the strongest prior attack.
           The first block averages across datasets for each target model;
           the second block averages across target models for each dataset. Except for the shaded row, all are zero‑shot cross-model evaluation. We include complete results in Appendix D. Our model outperforms all baselines significantly in all settings.}
  \label{tab:summary_table}
\end{table}

\subsection{Experimental Results}
\textbf{LAMP outperforms by a significant margin across multi-image benchmarks and models.} LAMP is compared with other baselines e.g.\ CPGC-UAP \cite{cpgc}, UAP-VLP \cite{uap_vlp}, Doubly-UAP \cite{Doubly-UAP}, Jailbreak-MLLM \cite{jailbreak-mllm}. Here, the last two baselines are multimodal baselines, and the first two are encoder-decoder baselines. Additionally, we also compared our method with other transferability-based methods like \cite{pandoras_box, mf-attack}. These methods are designed to learn universal adversarial perturbations and can be directly adapted to our problem. Other methods \cite{skip, adv_vit, transfer, boosting}  either fully rely on the output from classifiers or combine feature perturbation with classification loss \cite{enhance, inkawhichtransferable, perturbing}. Note that we learned UAPs based on the Mantis-CLIP pre-trained model, and the learned UAPs are applied across target MLLMs. LAMP achieves  $19.5\%$ average ASR gain across all models and datasets, as shown in Tab. \ref{tab:summary_table}. The specific per-model, per-dataset results are shown in Appendix D. Here, the optimal number of perturbations is $|\delta| = 2$.
\noindent\textbf{LAMP outperforms single image and multi-image VQA tasks.} In Tab.~\ref{tab:single_performance}, LAMP outperforms baselines by a significant margin on both single-image tasks such as LLaVA Bench and MM-Vet, as well as multi-image VQA tasks like Mantis Eval. We also present additional selection free VQA and image captioning task result in Appendix H 

\noindent\textbf{Ablation over loss components}. We have evaluated the different combination of the loss function Eq. \ref{eq:final_loss} in Tab. \ref{tab:loss_ablations}. If we skip $\mathcal{L}_{adv}^{ctg}$ and $\mathcal{L}_{adv}^{ias}$, the performance of LAMP drops.

\noindent\textbf{Robustness to defense strategy.} Following \cite{pandoras_box}, we evaluate the robustness of our attack method against defense mechanisms designed for different threat models. PatchCleanser \cite{patchcleanser} is a certifiable defense against adversarial patch attacks that uses a double masking strategy to certify predictions. As our method does not depend on visible patches, PatchCleanser does not apply to imperceptible attacks like ours. Instead, we evaluate against query-based defenses, which specifically detect malicious queries in black-box settings. Tab. \ref{tab:defense_robustness} shows that our attack remains robust in the presence defenses.

\noindent\textbf{Complexity analysis and hyperparameter sensitivity.} The attack complexity and hyperparameters ablations are shown in Appendix I and J.

\begin{table}[]
    \centering
    \begin{tabular}{c|c|c}
    \toprule
         Defense & Method &  ASR \\
         \midrule
         \citet{random_noise}    & LAMP &  70.23\%\\
        \citet{blacklight}  & LAMP  & 69.21\%\\
         \bottomrule
     \citet{random_noise}    &  \citet{pandoras_box}   &  56.33\%\\
         \citet{blacklight}    & \citet{pandoras_box}  & 20.21\%\\
        \bottomrule
    \end{tabular}
    \caption{ASR against blackbox defense strategies on Mantis Eval dataset and Mantis-CLIP model. Detailed defense results are provided in Appendix G.}
    \label{tab:defense_robustness}
\end{table}

% \noindent\textbf{Impact of ``contagious" and index-attention loss}
% In Tab. \ref{tab:loss_ablations}, we report the ASR for different combinations of loss functions and examine how ablating specific losses impacts the performance of LAMP. We observe that omitting the $\mathcal{L}_{adv}^{ias}$ and $\mathcal{L}_{adv}^{ctg}$  loss from LAMP leads to a significant drop in performance compared to other loss functions. We demonstrate that the impact of "contagious" objective in attention weights of input tokens in  \ref{sec:ctg_impact}. 

\noindent

\begin{table}[!ht]
\centering
\resizebox{\columnwidth}{!}{%
\begin{tabular}{c|c|c|c|c|c|c|c}
\toprule
 \multicolumn{5}{c|}{Loss}  &  \multicolumn{3}{c}{Datasets} \\
\cmidrule{1-7}

$\mathcal{L}_{adv}^{lm}$ &  $\mathcal{L}_{\text{adv}}^{dec}$ & $\mathcal{L}_{\text{adv}}^{h}$ & $\mathcal{L}_{adv}^{ctg}$  & $\mathcal{L}_{adv}^{ias}$ & Mantis Eval & NLVR2 & BLINK \\
\midrule

% \ding{51}& \ding{51} &  \ding{51} & \ding{51} & \ding{51}  & 73.43 & 52.73 & 84.86 \\
% &  \ding{51}& \ding{51}&  \ding{51}&  &67.12 & 48.56 & 78.90 \\
% \ding{51}& & \ding{51} & \ding{51} & &67.89 & 48.10 & 77.87 \\
% \ding{51}& \ding{51} & & \ding{51} & &67.32 & 48.34 & 78.90 \\
% \ding{51} & \ding{51} & \ding{51} & & &68.66 & 44.35 & 74.34 \\

\ding{51}& \ding{51} &  \ding{51} & \ding{51} & \ding{51}  & 73.43 & 52.73 & 84.86 \\
\ding{51}& \ding{51} &  \ding{51} &  & \ding{51}  & 70.32 & 49.33 & 82.64 \\
&  \ding{51}& \ding{51}&  \ding{51}&  &67.12 & 48.56 & 78.90 \\
\ding{51}& & \ding{51} & \ding{51} & &67.89 & 48.10 & 77.87 \\
\ding{51}& \ding{51} & & \ding{51} & &67.32 & 48.34 & 78.90 \\
\ding{51} & \ding{51} & \ding{51} & & &68.66 & 44.35 & 74.34 \\

\bottomrule
  
\end{tabular}%
}
\caption{ASR(\%) on three benchmark tasks with different combination of loss and comparison with LAMP.} 
\label{tab:loss_ablations}
\end{table}

%%% VQA tasks%%%
\begin{table}[ht!]
\centering
% \scriptsize
\resizebox{\columnwidth}{!}{%
\setlength{\tabcolsep}{1mm}
\begin{tabular}{c| c| c| c|c}
\toprule
\textbf{Target Model} & \textbf{Method} & \textbf{Mantis Eval} & \textbf{MM-Vet}  & \textbf{LLaVA Bench} \\ 
\midrule
\multirow{4}{*}{\centering  Mantis-CLIP} & CPGC-UAP      & 49.23 & 50.32 & 45.60 \\  
           & Jailbreak-MLLM       & 44.45 & 48.21 & 35.67 \\
 &   Doubly-UAP & 46.45 & 50.21 & 37.67 \\       
           & LAMP           & 70.32 & 73.45 & 68.31 \\ 
\midrule

\multirow{4}{*}{\centering VILA - 1.5} & CPGC-UAP      & 49.23 & 48.27 & 43.76 \\  
           & Jailbreak-MLLM & 15.56 & 17.24 & 18.65 \\ 
&   Doubly-UAP  & 45.45 & 49.13 & 38.74 \\  
           & LAMP           & 71.32 & 72.54 & 67.13 \\ 
\midrule

\multirow{4}{*}{\centering MiniGPT4} & CPGC-UAP        & 45.23 & 44.75 & 42.64 \\  
           & Jailbreak-MLLM & 13.56 & 12.24 & 13.65 \\ 
           &   Doubly-UAP  & 43.34 & 47.67 & 36.46 \\  
           & LAMP           & 69.23 & 68.54 & 66.13 \\ 

\bottomrule
\end{tabular}
}
\caption{Performance comparison on benchmark datasets \\for single and multi-image VQA tasks.}
\label{tab:single_performance}
\end{table}

\begin{figure*}[ht]
\centering
% \vspace{-1em} %% need to adjust if not fit
\begin{minipage}[t]{\textwidth}
\includegraphics[width=\linewidth]{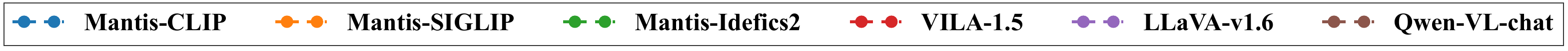}
\end{minipage} %\hfill
\begin{minipage}[t]{0.47\columnwidth}
  \includegraphics[width=\linewidth]{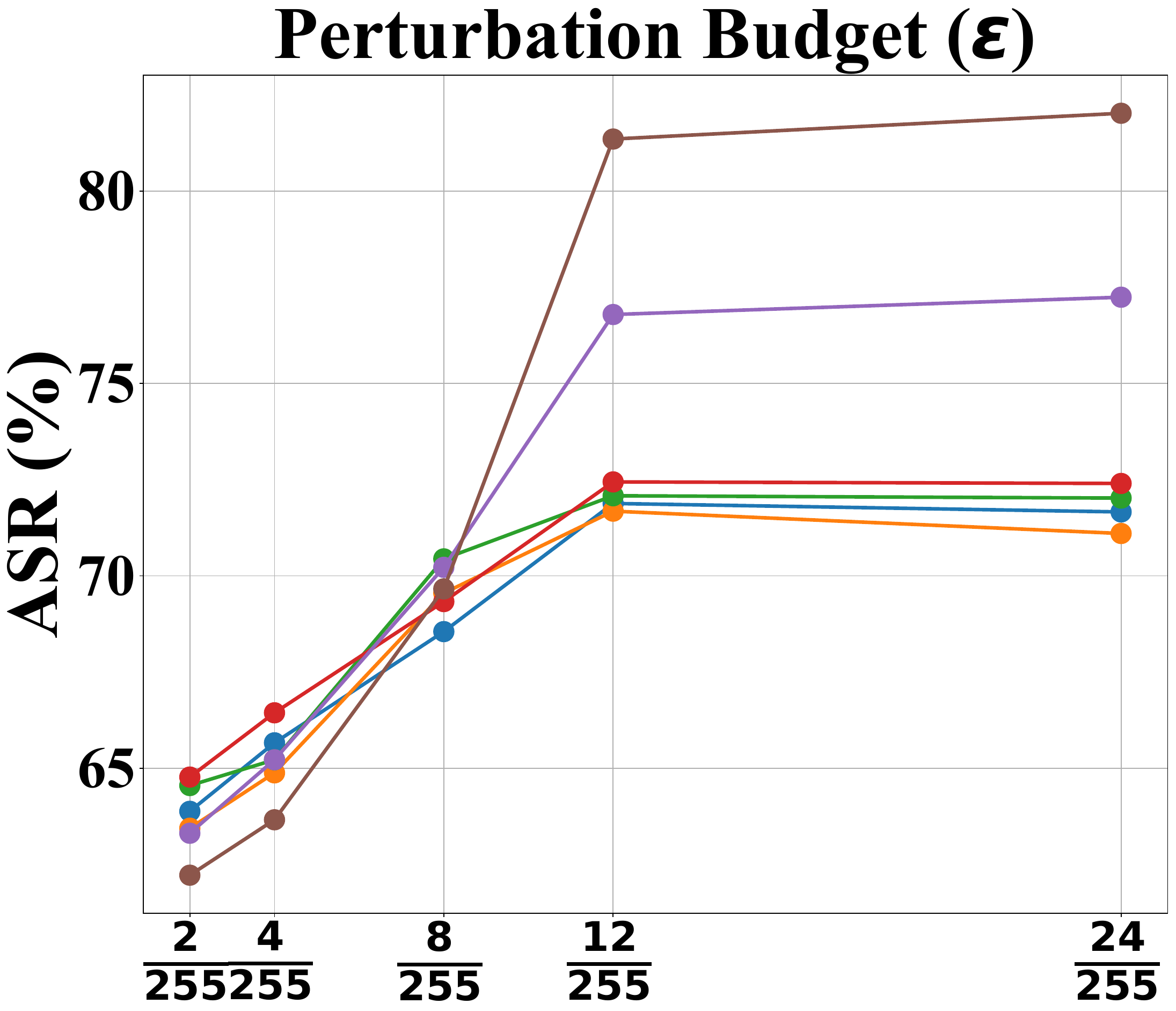}
  \subcaption{Perturbation budget vs ASR}
   \label{fig:budget}
\end{minipage} %\hfill 
\begin{minipage}[t]{0.47\columnwidth}
  \includegraphics[width=\linewidth]{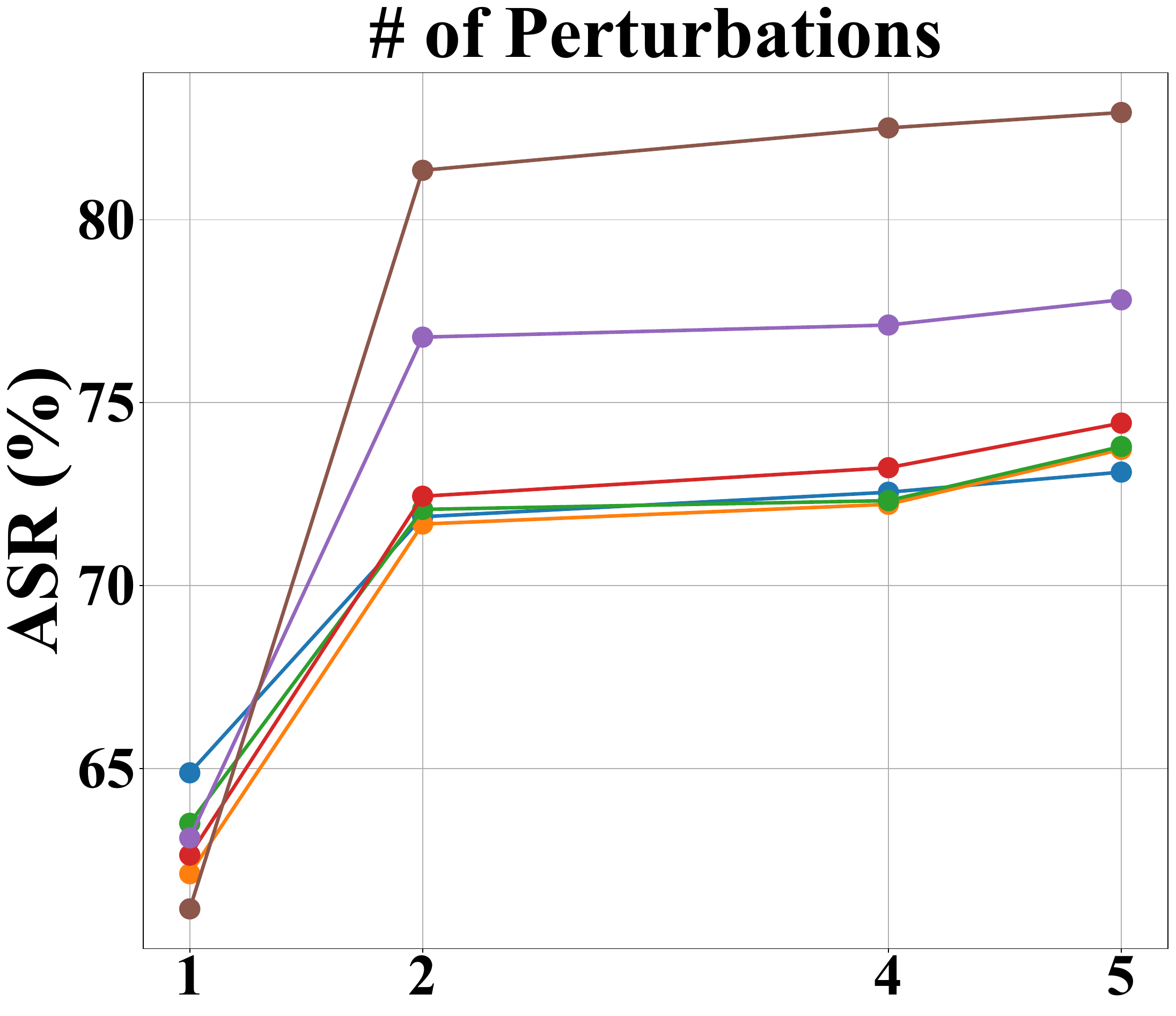}
  \subcaption{\# of perturbation vs ASR}
  \label{fig:perturb}
\end{minipage} %\hfill 
\begin{minipage}[t]{0.47\columnwidth}
  \includegraphics[width=\linewidth]{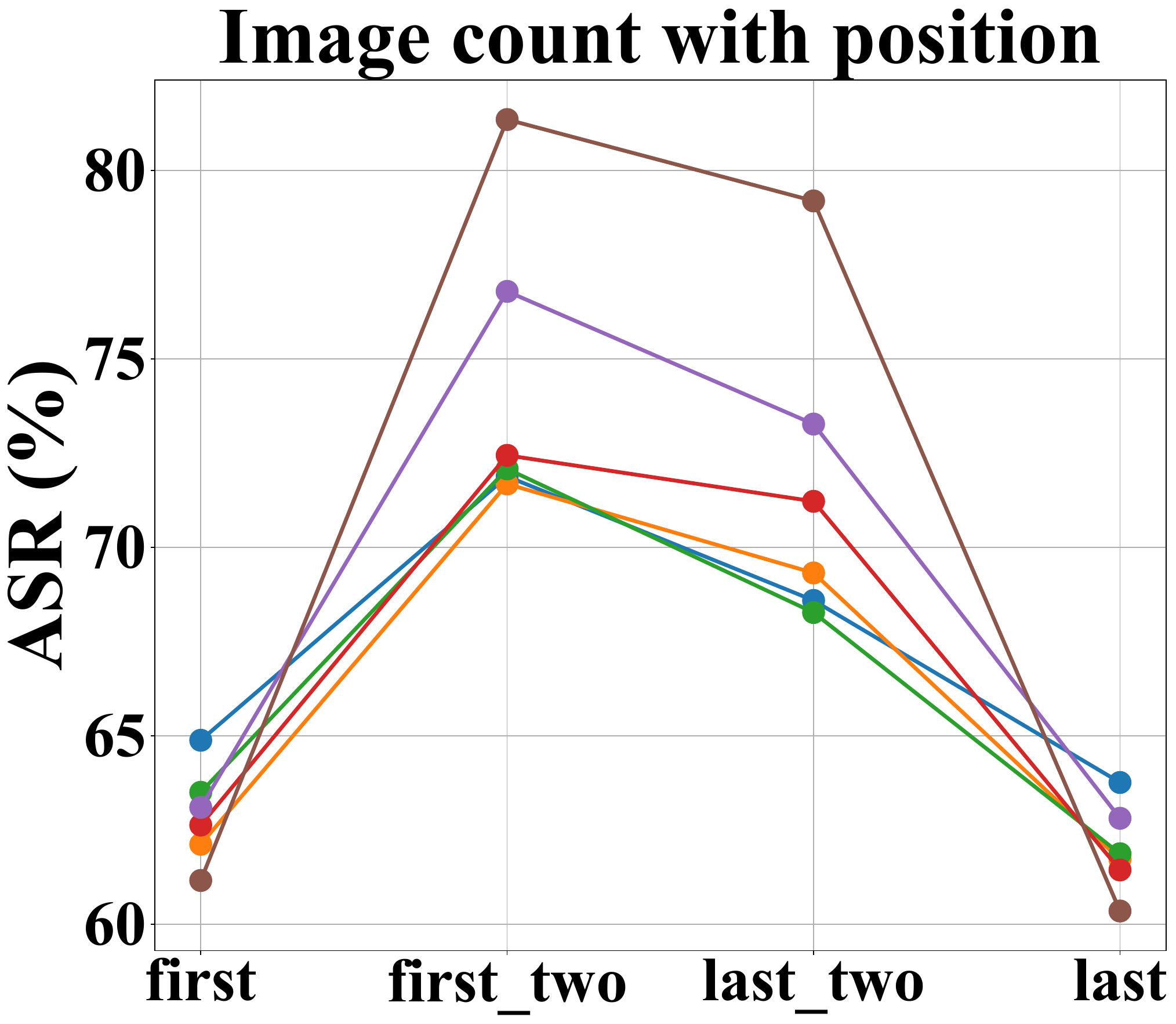}
  \subcaption{Image count vs ASR}
   \label{fig:image_pos}
\end{minipage} %\hfill 
\begin{minipage}[t]{0.47\columnwidth}
     \includegraphics[width=\linewidth]{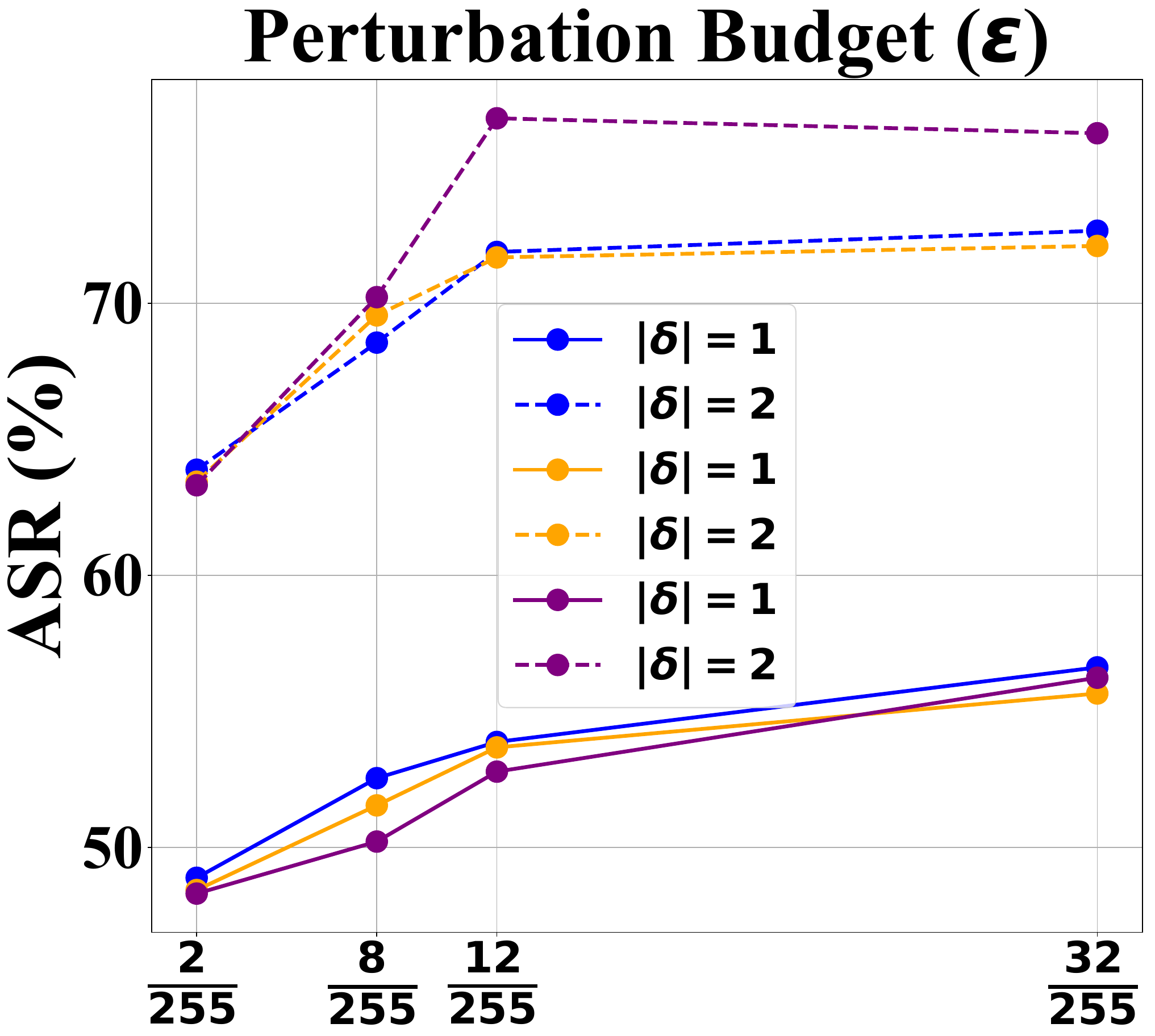}
  \subcaption{\# Perturbation vs. budget}
  \label{fig:compare_perturb_count}
\end{minipage} %\hfill 
\label{fig:ablations}
\caption{Impact of different hyperparameters in ASR.}
\end{figure*}

% \begin{figure}[htb]
%     \centering
%     \includegraphics[width=\linewidth]{LaTeX/figures/ablations_lamp_crop.pdf}

% \caption{Impact of different hyperparameters in ASR.}
% \label{fig:ablations}
% \end{figure}

% \vspace{-1em}
\subsection{Ablations}
\textbf{Perturbation budget vs ASR.} We compare the performance of ASR for different perturbation budget in Fig. \ref{fig:budget}. Here, the method with \textbf{``contagious"} attack. We observe that ASR improves significantly with the increasing perturbation budget. We experiment with $\epsilon= 12/255$ for imperceptibility, and increasing this value does not significantly improve ASR, but it compromises imperceptibility.

\noindent\textbf{\# of perturbations vs ASR.} We compare the performance of ASR for different number of perturbations for Mantis-CLIP and Mantis-Eval datasets with \textbf{``contagious"} attack Fig. \ref{fig:perturb}. We observe that ASR improves significantly when the number of universal perturbations goes from $1$ to $2$, but after that, it does not improve significantly. We argue that, the ``contagious" attack impacts the clean images in attention spaces that help us to gain the similar ASR even number of perturbations more than $2$.

\noindent\textbf{Perturbation position vs ASR. } We compare the performance of ASR with different number of perturbations at different position of interleaved image-text inputs in Fig. \ref{fig:image_pos}. When the front two images are perturbed the ASR is the best, the ASR decreases a little as the last two images are perturbed, and least ASR for first and last image.

\noindent\textbf{Comparing perturbation count vs. budget} In Fig. \ref{fig:compare_perturb_count}, when $|\delta|=2$, we achieve a very high ASR compared to when $|\delta|=1$ for all perturbation budget. We can also infer that if  $|\delta|=2$, we can maintain a very ASR in low budget maintaining the imperceptibility. \noindent\textbf{Quantification of imperceptibility.} To evaluate the imperceptibility of the adversarial perturbations, we adopt perceptual similarity metrics, including LPIPS, as detailed in Appendix K. Since lower LPIPS values correspond to more imperceptibility, our method (0.021) demonstrates significantly improved stealth compared to the best-performing baseline (0.068).
\noindent\textbf{Interaction between losses.} We present interaction between losses and analysis of "contagious" losses in Appendix L.

% \textbf{Epochs vs ASR.} We compare the ASR performance at different epochs Fig. \ref{fig:epoch_asr}. 

% \input{AnonymousSubmission/LaTeX/sections/qual}
\section{Qualitative Analysis}
\label{sec:qual}
\begin{figure}[htb]
    \centering
    \includegraphics[width=\linewidth]{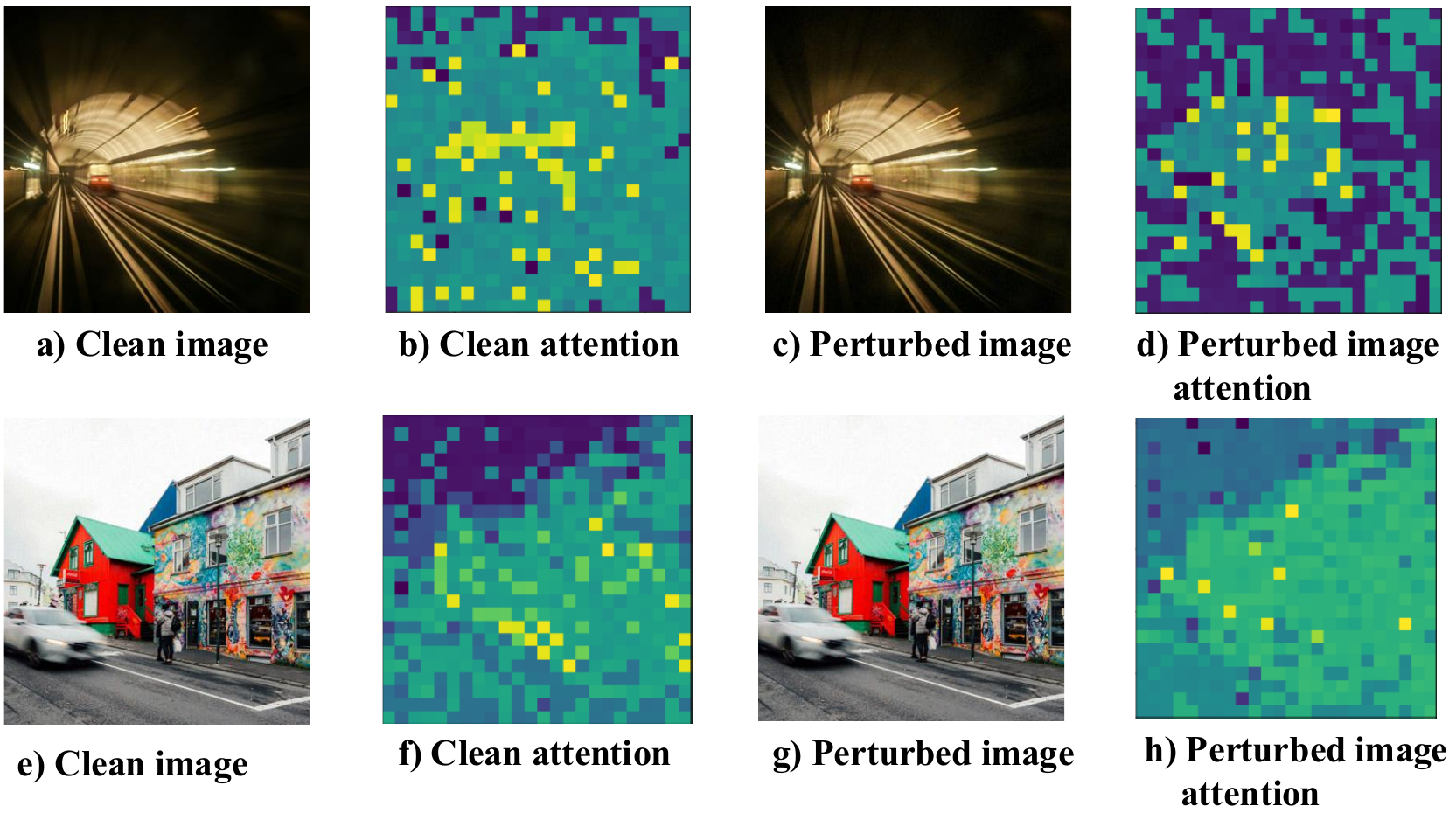}
    How many running white compact cars are there in all images? (A) One \quad (B) Two \quad (C) Three. \\
Answer with the option's letter from the given choices directly. GT: A, Output: \textcolor{red}{B}

\caption{Attention maps of clean and perturbed images. 
Here, two input images and one question. The incorrect answer is in red. Yellow indicates high attention.}
\label{fig:attention_map}
\end{figure}

 We visualize the effect of our proposed method through attention maps in Fig. \ref{fig:attention_map}. For clean images, the downstream model pays attention to object of the images (distribution of the yellow dots). However, when the imperceptible perturbations are added to the images, the model starts to pay attention to a different location of the images. We have also shown position- invariant attack in Appendix E.

\section{Conclusion}
\label{sec:conclusion}

 In this paper, we investigate to learn UAPs that are capable of transferring across different target multi-image MLLMs
models, datasets and downstream tasks. We propose a novel UAP 
learning method LAMP that incorporates different constraints exploiting the self attention module of the LLM backbone. We propose a novel "contagious" constraint that enables an attacker to learn perturbation by infecting the clean tokens through self attention. We also propose an index-attention suppression objective so that the attack remains position-invariant.  
We test the proposed methods across different
target MLLMs, downstream tasks, and promising results demonstrate the superiority of the proposed method.

\section{Acknowledgments}
This work was supported in part by a Google Research Scholar award and Virginia Commonwealth Cyber Initiative Award \#469112. We acknowledge Advanced Research Computing (ARC) at Virginia Tech for providing the computational resources and technical support that contributed to the results reported in this paper. The authors would also like to thank the reviewers for their constructive feedback.

\input{appendix}

\bibliographystyle{plainnat}   % or IEEEtran, unsrt, etc.

\input{Formatting-Instructions-LaTeX-2026.bbl}
\end{document}

%% file: appendix.tex
\setcounter{secnumdepth}{1} %May be changed to 1 or 2 if section numbers are desired.

% The file aaai2026.sty is the style file for AAAI Press

% Title

% Your title must be in mixed case, not sentence case.
% That means all verbs (including short verbs like be, is, using,and go),
% nouns, adverbs, adjectives should be capitalized, including both words in hyphenated terms, while
% articles, conjunctions, and prepositions are lower case unless they
% directly follow a colon or long dash

% \section{Appendix}
\appendix
\label{sec:appendix}
% \section{Appendix}
\section{Impact of contagious  objective in attention weights}
\label{sec:ctg_impact}

\begin{figure*}[ht]
     \centering
 \begin{minipage}[t]{\columnwidth}
  \includegraphics[width=\linewidth]
  {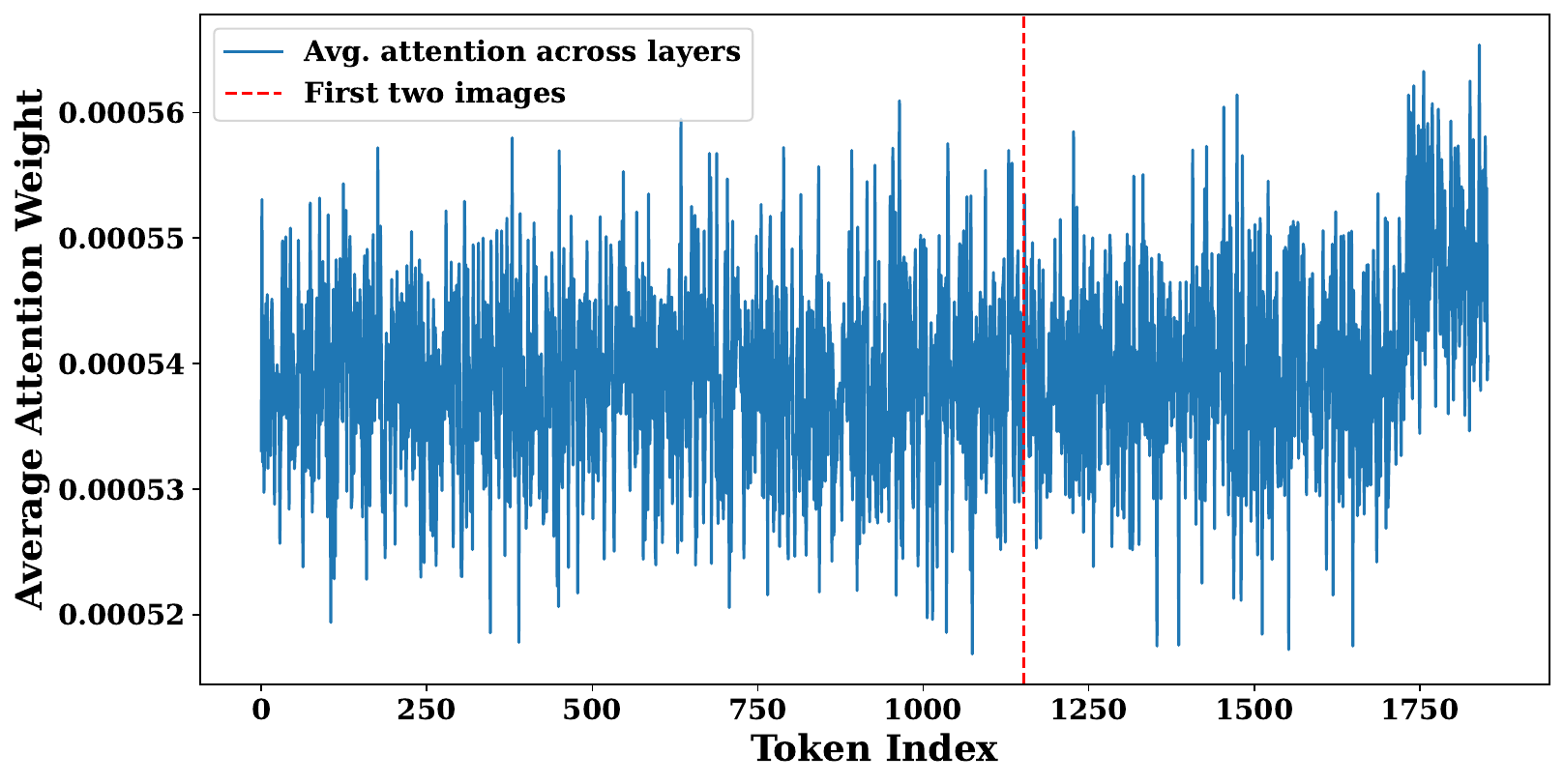}
  \subcaption{Average attention weights of input tokens across LLM}
   \label{fig:clean_attn}
\end{minipage} %\hfill 
\begin{minipage}[t]{\columnwidth}
  \includegraphics[width=\linewidth]{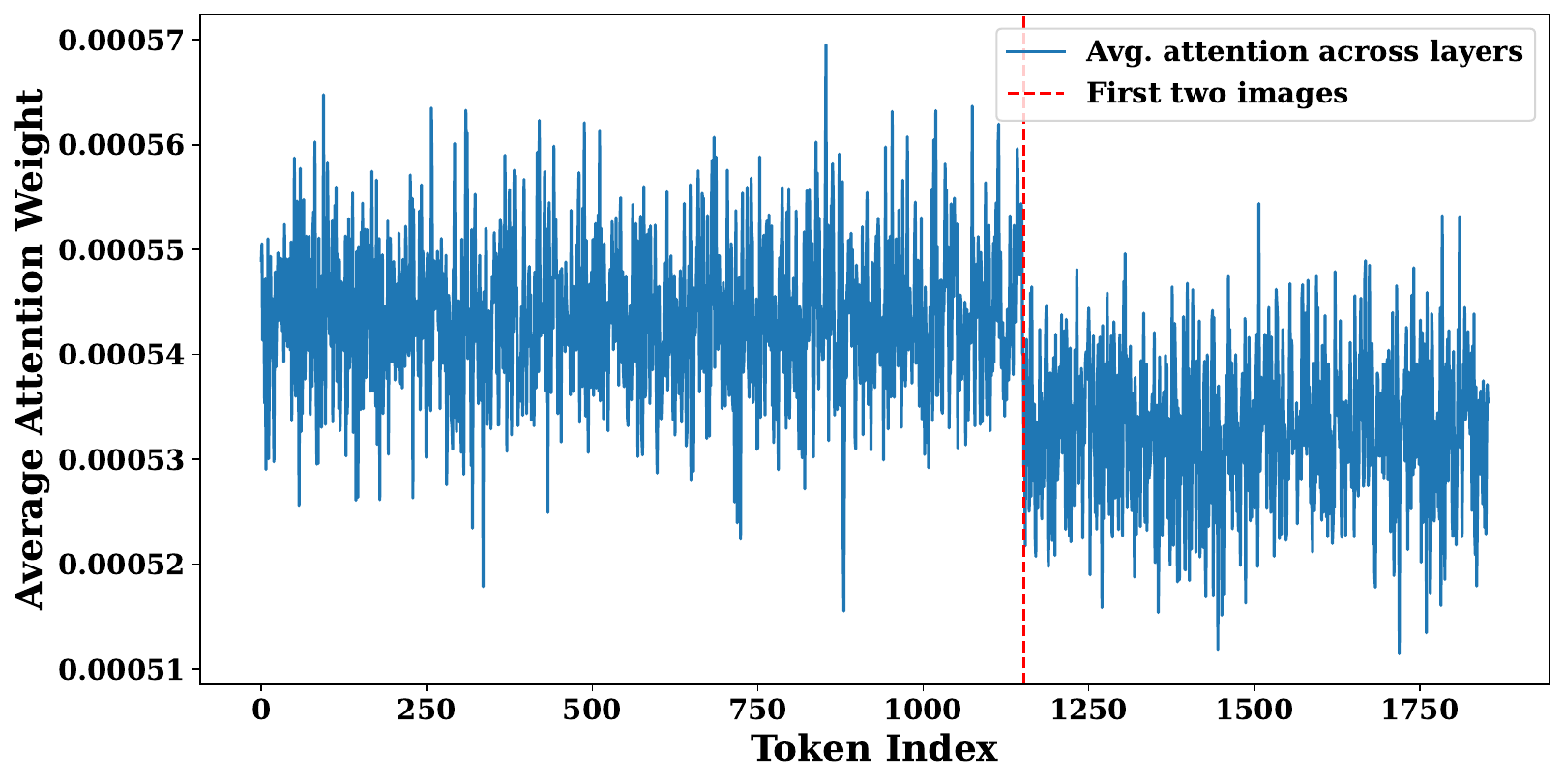}
  \subcaption{Average attention weights of perturbed input tokens across LLM for "contagious" objective.}
  \label{fig:ctg_attn}
\end{minipage} %\hfill 
\caption{Impact of contagious objective}
\label{fig:compare_ctg}
\end{figure*}

In Fig. \ref{fig:clean_attn}, we show the average attention weights across decoder layers for clean inputs. We observe that the LLM pays more attention to later tokens, most likely text tokens. Next, we learn UAPs using only the contagious objective. These UAPs are added to the first two images of a sample. In Fig. \ref{fig:ctg_attn}, we show the average attention for the same sample after the first two images are perturbed with the learned UAPs. Based on the contagious objective, the clean tokens pay more attention to the perturbed tokens (i.e., the tokens corresponding to the first two images), which validates the concept of the contagious loss.

\begin{figure}[h!]
    \centering
    \includegraphics[width=\linewidth, height=5.5cm]{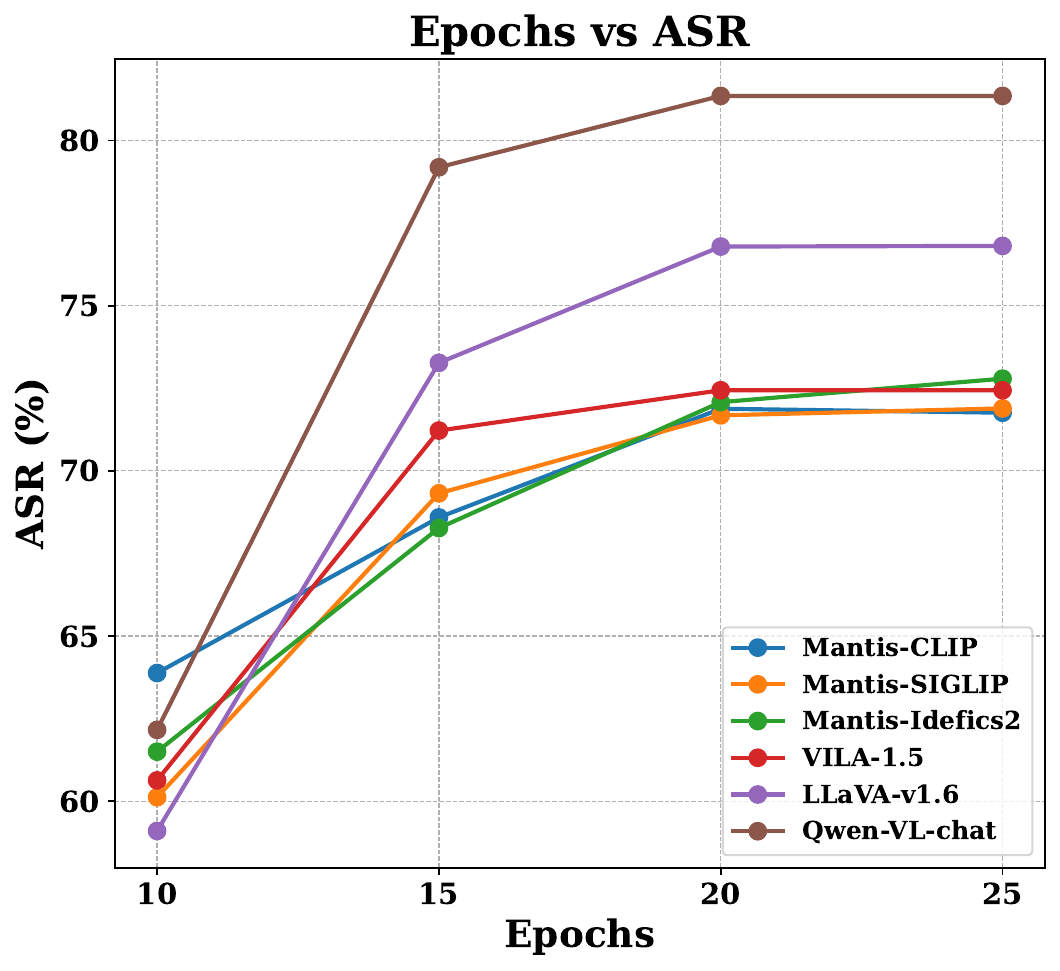}
    \caption{ASR vs Epoch with UAP learning in LAMP}
    \label{fig:ASRvsEpoch}
\end{figure}

\begin{table}[hbt!]
    \centering
    \begin{tabular}{c|c}
    \toprule
        Datasets & \# Instances \\
        \midrule
        NLVR2 & 6967 \\
        Mantis-Eval & 217 \\
        BLINK & 1901 \\
        Q-bench & 1000 \\
        MVBench & 4000\\
    \end{tabular}
    \caption{Dataset statistics.}
    \label{tab:dataset_stats}
\end{table}

\vspace{-1em}
\section{ASR vs Epoch}
In Fig. \ref{fig:ASRvsEpoch}, we show the ASR performance variation across epochs. After epoch $20$, ASR does not change significantly. For this reason, we set the epoch to $20$ for learning UAPs in our standard experiments. 

\section{Benchmark datasets statistics}
\label{sec:dataset_stats}
In Tab. \ref{tab:dataset_stats}, the instances of multi-image benchmarks are reported.

\section{Comparison with baselines with LAMP and different combination of losses}
\label{sec:comparsion_lamp}

Tab. \ref{tab:main_table} shows the ASR of LAMP compared across benchmark datasets and baseline models. Here, the universal perturbations are learned from the Mantis-CLIP model, and zero-shot evaluation was done across other models to show generalization capability. We take the average best baseline across models and datasets and show the aggregated results in Tab.~\ref{tab:aux_table}. Compared to all baselines, we achieve significant performance gains, about $19.5\%$ on average across all models and datasets. We have also shown the ablation of LAMP (different combination of loss functions) with different baselines in Tab. \ref{tab:aux_table}. The table shows that when different parts of the loss functions are aggregated the ASR increases across models and datasets.

\begin{table*}[!ht]
\centering
\resizebox{\textwidth   }{!}{%
\begin{tabular}{c|c|c|c|c|c|c}
\toprule
\multirow{2}{*}{\centering \textbf{Target Model}} & \multirow{2}{*}{\centering \textbf{Method}} &  \multicolumn{5}{c}{\textbf{Benchmark Datasets}} \\
\cmidrule(lr){3-7}
& & Mantis Eval $\uparrow$ & NLVR2 $\uparrow$  & BLINK $\uparrow$   & Q-Bench $\uparrow$  & MVBench $\uparrow$ \\ 
% \cmidrule{1-7}
\midrule

 % &  Clean(No attack) & 55.76 & 84.66 & 47.06 & 66.00 &  48.30  \\
\multirow{6}{*}{\centering Mantis-CLIP} & CPGC-UAP \cite{cpgc}  & 51.66          & 19.33          & 56.44          & 39.33          & 54.35   \\
 & UAP-VLP \cite{uap_vlp} & 54.68          & 24.75          & 61.77          & 43.66          & 59.66   \\

 & Doubly-UAP \cite{Doubly-UAP} &  56.87          & 30.55          & 63.74    & 44.76          & 61.63   \\
 & Jailbreak-MLLM \cite{jailbreak-mllm} & 54.68          & 24.75          & 61.77          & 43.66          & 59.66   \\

  &Pandoras's box \cite{pandoras_box} & 57.58          & 26.55          & 65.52          & 46.64          & 60.61   \\

   &X-transfer \cite{x_transfer} & 52.82          & 22.43          & 62.24          & 40.87          & 57.11   \\
% & $\mathcal{L}_{adv}^{lm}$  & 59.55       & 30.66 & 64.68 & 52.79   & 61.79 \\
% & $\mathcal{L}_{adv}^{lm} +\mathcal{L}_{\text{adv}}^{dec}$ & 63.68 & 36.55 & 69.68 & 56.55 & 66.44  \\  

% & $\mathcal{L}_{adv}^{lm} +\mathcal{L}_{\text{adv}}^{dec} + \mathcal{L}_{\text{adv}}^{h}$ & 68.66 & 44.35 & 74.34 & 62.12 & 70.68 \\

% & $\mathcal{L}_{adv}^{lm} +\mathcal{L}_{\text{adv}}^{dec} + \mathcal{L}_{\text{adv}}^{h} + \mathcal{L}_{adv}^{ctg}$ & {71.88}          & {50.47}          & {80.86}          & {68.64}          & {75.55} \\ 

&\textbf{LAMP} & \textbf{73.43}          & \textbf{52.73}          & \textbf{84.46}          & \textbf{71.42}          & \textbf{77.67}    \\

\cmidrule{1-7}
\multicolumn{7}{c}{Zero-shot Cross-Model Evaluation} \\
\cmidrule{1-7}

% Model & Method &  \multicolumn{5}{c}{Benchmark Datasets} \\
% \cmidrule{1-7}
% & & Mantis Eval $\downarrow$ & NLVR2 $\downarrow$  & BLINK $\downarrow$   & Q-Bench $\downarrow$  & MVBench $\downarrow$ \\ 
% \cmidrule{3-7}
% &  Clean(No attack) & 59.45  & 87.43 & 46.35 & 69.90 &  50.15   \\
\multirow{6}{*}{\centering Mantis-SIGLIP} & CPGC-UAP \cite{cpgc} & 52.13 & 19.88 & 57.77 & 38.77 & 55.68  \\
 & UAP-VLP \cite{uap_vlp} & 55.55 & 25.44 & 60.44 & 42.55 & 59.68 \\

  & Doubly-UAP \cite{Doubly-UAP} &  56.74 & 31.43          & 62.34    & 43.64          & 63.63   \\
 & Jailbreak-MLLM \cite{Doubly-UAP} & 53.68          & 22.53          & 60.43          & 42.64 & 57.67   \\
 
% & $\mathcal{L}_{adv}^{lm}$   & 58.79 & 30.80 & 65.09 & 52.90 & 60.10    \\
% & $\mathcal{L}_{adv}^{lm} +\mathcal{L}_{\text{adv}}^{dec}$ & 62.18 & 36.79 & 70.02 & 56.44 & 66.79  \\

% & $\mathcal{L}_{adv}^{lm} +\mathcal{L}_{\text{adv}}^{dec} + \mathcal{L}_{\text{adv}}^{h}$ &  66.79 & 42.26 & 76.55 & 63.79 & 71.66\\

% & $\mathcal{L}_{adv}^{lm} +\mathcal{L}_{\text{adv}}^{dec} + \mathcal{L}_{\text{adv}}^{h} + \mathcal{L}_{adv}^{ctg}$  & {71.68} & {50.77} & {80.66} & 68.33 & 76.79   \\  

&\textbf{LAMP}& \textbf{73.21}          & \textbf{52.74}          & \textbf{83.65}          & \textbf{71.45}          & \textbf{78.57} \\ 

\cmidrule{1-7}

% &  Clean(No attack) & 57.14  & 89.71 & 49.05 & 75.20 &  51.38   \\
\multirow{6}{*}{\centering Mantis-Idefics2} & CPGC-UAP \cite{cpgc}  &  51.79 & 20.77 & 57.79 & 31.33 & 54.32 \\
 & UAP-VLP \cite{uap_vlp} & 54.22 &  23.11 & 60.24 & 34.66 & 60.24\\

 & Doubly-UAP \cite{Doubly-UAP} &  52.74 & 29.43 & 60.42    & 40.43          & 61.63   \\
 & Jailbreak-MLLM \cite{jailbreak-mllm} & 40.81          & 19.53          & 50.43          & 40.41 & 42.71   \\
 
% & $\mathcal{L}_{adv}^{lm}$   & 57.79       & 29.44 & 66.84 & 51.14   & 61.10     \\
% & $\mathcal{L}_{adv}^{lm} +\mathcal{L}_{\text{adv}}^{dec}$ &  61.79 & 39.66 & 74.44 & 57.48 & 67.55 \\

% & $\mathcal{L}_{adv}^{lm} +\mathcal{L}_{\text{adv}}^{dec} + \mathcal{L}_{\text{adv}}^{h}$ & 66.66 & 43.13 & 78.77 & 62.11 & 71.09 \\

% & $\mathcal{L}_{adv}^{lm} +\mathcal{L}_{\text{adv}}^{dec} + \mathcal{L}_{\text{adv}}^{h} + \mathcal{L}_{adv}^{ctg}$ & 72.08 & {51.55} & {82.77} & {68.78} & {73.68} \\

& \textbf{LAMP} & \textbf{75.44}          & \textbf{53.51}          & \textbf{85.79}          & \textbf{70.55}          & \textbf{76.78} \\ 

\cmidrule{1-7}

% &  Clean(No attack) & 51.15  & 76.45 & 39.30 & 45.70 &  49.40 \\
\multirow{6}{*}{\centering VILA-1.5} & CPGC-UAP \cite{cpgc} & 52.68 & 29.68 & 66.65 & 60.09 & 56.24   \\
 & UAP-VLP \cite{uap_vlp} & 54.79 & 33.22 & 65.44 & 65.33 & 57.79\\

 & Doubly-UAP \cite{Doubly-UAP} &  50.74 & 28.36 & 59.23    & 41.35          & 60.33   \\
 & Jailbreak-MLLM \cite{jailbreak-mllm} & 36.15          & 21.33          & 45.43          & 38.43 & 40.13   \\
 
% & $\mathcal{L}_{adv}^{lm}$  & 58.75 & 38.46 & 66.66 & 67.90 & 59.77   \\
% & $\mathcal{L}_{adv}^{lm} +\mathcal{L}_{\text{adv}}^{dec}$ &  62.12 & 44.57 & 71.57 & 75.33 & 63.44 \\

% & $\mathcal{L}_{adv}^{lm} +\mathcal{L}_{\text{adv}}^{dec} + \mathcal{L}_{\text{adv}}^{h}$ & 67.57 & 49.44 & 76.55 & 79.66 & 69.76 \\

% & $\mathcal{L}_{adv}^{lm} +\mathcal{L}_{\text{adv}}^{dec} + \mathcal{L}_{\text{adv}}^{h} + \mathcal{L}_{adv}^{ctg}$ & {72.44} & {56.79} & {81.10} & {84.74} & {74.10}  \\ 

& \textbf{LAMP} & \textbf{75.21}          & \textbf{57.43}          & \textbf{83.56}          & \textbf{87.23}          & \textbf{77.81} \\

\cmidrule{1-7}
% & Clean(No attack) & 45.62   & 58.88 & 39.55 & 54.80 &  40.90 \\
\multirow{6}{*}{\centering LLaVa-v1.6} & CPGC-UAP \cite{cpgc} & 59.76 & 48.74 & 67.55 & 49.09 & 63.35     \\
 & UAP-VLP \cite{uap_vlp}  & 60.79 & 48.66 & 68.66 & 50.10 & 64.44\\
 
 & Doubly-UAP \cite{Doubly-UAP} &  51.43 & 29.64 & 52.34    & 42.53          & 55.31   \\
 & Jailbreak-MLLM \cite{jailbreak-mllm} & 36.15          & 21.33          & 45.43          & 38.43 & 40.13   \\
 
% & $\mathcal{L}_{adv}^{lm}$  & 64.76 & 54.79 & 72.78 & 55.44 & 67.77   \\
% & $\mathcal{L}_{adv}^{lm} +\mathcal{L}_{\text{adv}}^{dec}$ & 66.68 & 57.44 & 75.15 & 59.76 & 69.88 \\

% & $\mathcal{L}_{adv}^{lm} +\mathcal{L}_{\text{adv}}^{dec} + \mathcal{L}_{\text{adv}}^{h}$ & 71.47 & 59.65 & 79.55 & 67.66 & 74.22  \\

% & $\mathcal{L}_{adv}^{lm} +\mathcal{L}_{\text{adv}}^{dec} + \mathcal{L}_{\text{adv}}^{h} + \mathcal{L}_{adv}^{ctg}$  & {76.79} & {64.22} & {84.33} & {74.55} & {79.24}  \\ 

& \textbf{LAMP} & \textbf{79.65}          & \textbf{67.51}          & \textbf{86.79}          & \textbf{77.55}          & \textbf{82.78} \\

\cmidrule{1-7}

% & Clean(No attack) & 39.17   & 58.72 & 31.17 & 45.90 &  42.15 \\
\multirow{6}{*}{\centering Qwen-VL-Chat} & CPGC-UAP \cite{cpgc} & 64.68 & 49.68 & 72.11 & 59.74 & 61.09   \\
 & UAP-VLP \cite{uap_vlp} & 66.68 & 52.11 & 74.22 & 63.29 & 65.55 \\

 & Doubly-UAP \cite{Doubly-UAP} &  49.42 & 25.44 & 50.43    & 40.51          & 53.31   \\
 & Jailbreak-MLLM \cite{jailbreak-mllm} & 33.15          & 20.33          & 43.43          & 33.34 & 38.32   \\
 
% & $\mathcal{L}_{adv}^{lm}$   & 69.66       & 56.57 & 76.66 & 67.44   & 68.22  \\
% & $\mathcal{L}_{adv}^{lm} +\mathcal{L}_{\text{adv}}^{dec}$ & 72.11 & 59.57 & 79.79 & 70.59 & 71.44  \\

% & $\mathcal{L}_{adv}^{lm} +\mathcal{L}_{\text{adv}}^{dec} + \mathcal{L}_{\text{adv}}^{h}$ & 77.55 & 62.55 & 81.08 & 72.13 & 75.68 \\

% & $\mathcal{L}_{adv}^{lm} +\mathcal{L}_{\text{adv}}^{dec} + \mathcal{L}_{\text{adv}}^{h} + \mathcal{L}_{adv}^{ctg}$ & {81.35} & {64.63} & {84.68} & {74.66}  & {79.55}    \\ 

& \textbf{LAMP} & \textbf{84.57}          & \textbf{67.15}          & \textbf{87.65}          & \textbf{76.12}          & \textbf{83.84} \\

\cmidrule{1-7}

\multirow{6}{*}{\centering Qwen-2.5} & CPGC-UAP \cite{cpgc} & 63.84 & 48.83 & 71.14 & 58.41 & 60.94   \\
 & UAP-VLP \cite{uap_vlp} & 65.45 & 50.15 & 72.24 & 60.23 & 64.59 \\

 & Doubly-UAP \cite{Doubly-UAP} &  48.41 & 23.42 & 48.43    & 38.13          & 50.31   \\
 & Jailbreak-MLLM \cite{jailbreak-mllm} & 30.76  & 19.54          & 41.34          & 30.43 & 36.23   \\
 
& \textbf{LAMP} & \textbf{82.71}          & \textbf{66.64}          & \textbf{87.65}          & \textbf{77.56} & \textbf{82.43} \\
\bottomrule

\end{tabular}%
}
\caption{Attack success rate (\%) on the multi-image benchmarks compared with baselines and variants of our method. $\mathcal{L}_{adv}^{lm}$,  $\mathcal{L}_{adv}^{lm} +\mathcal{L}_{\text{adv}}^{dec}$,    $\mathcal{L}_{adv}^{lm} +\mathcal{L}_{\text{adv}}^{dec} + \mathcal{L}_{\text{adv}}^{h}$ are our baselines.  Bold indicates the best method ("LAMP")}
\label{tab:main_table}
\end{table*}

%%%% additional losses

% \subsection{Comparison with baselines with different combination of losses}
% \label{sec:comparsion_baseline}

\begin{table*}[!ht]
\centering
\resizebox{\textwidth   }{!}{%
\begin{tabular}{c|c|c|c|c|c|c}
\toprule
\multirow{2}{*}{\centering \textbf{Target Model}} & \multirow{2}{*}{\centering \textbf{Method}} &  \multicolumn{5}{c}{\textbf{Benchmark Datasets}} \\
\cmidrule(lr){3-7}
& & Mantis Eval $\uparrow$ & NLVR2 $\uparrow$  & BLINK $\uparrow$   & Q-Bench $\uparrow$  & MVBench $\uparrow$ \\ 
% \cmidrule{1-7}
\midrule

 % &  Clean(No attack) & 55.76 & 84.66 & 47.06 & 66.00 &  48.30  \\
\multirow{6}{*}{\centering Mantis-CLIP} & CPGC-UAP \cite{cpgc}  & 51.66          & 19.33          & 56.44          & 39.33          & 54.35   \\
 & UAP-VLP \cite{uap_vlp} & 54.68          & 24.75          & 61.77          & 43.66          & 59.66   \\

 & Doubly-UAP &  56.87          & 30.55          & 63.74    & 44.76          & 61.63   \\
 & Jailbreak-MLLM & 54.68          & 24.75          & 61.77          & 43.66          & 59.66   \\
 
& $\mathcal{L}_{adv}^{lm}$  & 59.55       & 30.66 & 64.68 & 52.79   & 61.79 \\
& $\mathcal{L}_{adv}^{lm} +\mathcal{L}_{\text{adv}}^{dec}$ & 63.68 & 36.55 & 69.68 & 56.55 & 66.44  \\  

& $\mathcal{L}_{adv}^{lm} +\mathcal{L}_{\text{adv}}^{dec} + \mathcal{L}_{\text{adv}}^{h}$ & 68.66 & 44.35 & 74.34 & 62.12 & 70.68 \\

& $\mathcal{L}_{adv}^{lm} +\mathcal{L}_{\text{adv}}^{dec} + \mathcal{L}_{\text{adv}}^{h} + \mathcal{L}_{adv}^{ctg}$ & {71.88}          & {50.47}          & {80.86}          & {68.64}          & {75.55} \\ 

% &\textbf{LAMP} & \textbf{73.43}          & \textbf{52.73}          & \textbf{84.46}          & \textbf{71.42}          & \textbf{77.67}    \\  

\cmidrule{1-7}
\multicolumn{7}{c}{Zero-shot Cross-Model Evaluation} \\
\cmidrule{1-7}

% Model & Method &  \multicolumn{5}{c}{Benchmark Datasets} \\
% \cmidrule{1-7}
% & & Mantis Eval $\downarrow$ & NLVR2 $\downarrow$  & BLINK $\downarrow$   & Q-Bench $\downarrow$  & MVBench $\downarrow$ \\ 
% \cmidrule{3-7}
% &  Clean(No attack) & 59.45  & 87.43 & 46.35 & 69.90 &  50.15   \\
\multirow{6}{*}{\centering Mantis-SIGLIP} & CPGC-UAP \cite{cpgc} & 52.13 & 19.88 & 57.77 & 38.77 & 55.68  \\
 & UAP-VLP \cite{uap_vlp} & 55.55 & 25.44 & 60.44 & 42.55 & 59.68 \\

  & Doubly-UAP &  56.74 & 31.43          & 62.34    & 43.64          & 63.63   \\
 & Jailbreak-MLLM & 53.68          & 22.53          & 60.43          & 42.64 & 57.67   \\
 
& $\mathcal{L}_{adv}^{lm}$   & 58.79 & 30.80 & 65.09 & 52.90 & 60.10    \\
& $\mathcal{L}_{adv}^{lm} +\mathcal{L}_{\text{adv}}^{dec}$ & 62.18 & 36.79 & 70.02 & 56.44 & 66.79  \\

& $\mathcal{L}_{adv}^{lm} +\mathcal{L}_{\text{adv}}^{dec} + \mathcal{L}_{\text{adv}}^{h}$ &  66.79 & 42.26 & 76.55 & 63.79 & 71.66\\

& $\mathcal{L}_{adv}^{lm} +\mathcal{L}_{\text{adv}}^{dec} + \mathcal{L}_{\text{adv}}^{h} + \mathcal{L}_{adv}^{ctg}$  & {71.68} & {50.77} & {80.66} & 68.33 & 76.79   \\

% &\textbf{LAMP}& \textbf{73.21}          & \textbf{52.74}          & \textbf{83.65}          & \textbf{71.45}          & \textbf{78.57} \\ 

\cmidrule{1-7}

% &  Clean(No attack) & 57.14  & 89.71 & 49.05 & 75.20 &  51.38   \\
\multirow{6}{*}{\centering Mantis-Idefics2} & CPGC-UAP \cite{cpgc}  &  51.79 & 20.77 & 57.79 & 31.33 & 54.32 \\
 & UAP-VLP \cite{uap_vlp} & 54.22 &  23.11 & 60.24 & 34.66 & 60.24\\

 & Doubly-UAP &  52.74 & 29.43 & 60.42    & 40.43          & 61.63   \\
 & Jailbreak-MLLM & 40.81          & 19.53          & 50.43          & 40.41 & 42.71   \\
 
& $\mathcal{L}_{adv}^{lm}$   & 57.79       & 29.44 & 66.84 & 51.14   & 61.10     \\
& $\mathcal{L}_{adv}^{lm} +\mathcal{L}_{\text{adv}}^{dec}$ &  61.79 & 39.66 & 74.44 & 57.48 & 67.55 \\

& $\mathcal{L}_{adv}^{lm} +\mathcal{L}_{\text{adv}}^{dec} + \mathcal{L}_{\text{adv}}^{h}$ & 66.66 & 43.13 & 78.77 & 62.11 & 71.09 \\

& $\mathcal{L}_{adv}^{lm} +\mathcal{L}_{\text{adv}}^{dec} + \mathcal{L}_{\text{adv}}^{h} + \mathcal{L}_{adv}^{ctg}$ & 72.08 & {51.55} & {82.77} & {68.78} & {73.68} \\

% & \textbf{LAMP} & \textbf{75.44}          & \textbf{53.51}          & \textbf{85.79}          & \textbf{70.55}          & \textbf{76.78} \\ 

\cmidrule{1-7}

% &  Clean(No attack) & 51.15  & 76.45 & 39.30 & 45.70 &  49.40 \\
\multirow{6}{*}{\centering VILA-1.5} & CPGC-UAP \cite{cpgc} & 52.68 & 29.68 & 66.65 & 60.09 & 56.24   \\
 & UAP-VLP \cite{uap_vlp} & 54.79 & 33.22 & 65.44 & 65.33 & 57.79\\

 & Doubly-UAP &  50.74 & 28.36 & 59.23    & 41.35          & 60.33   \\
 & Jailbreak-MLLM & 36.15          & 21.33          & 45.43          & 38.43 & 40.13   \\
 
& $\mathcal{L}_{adv}^{lm}$  & 58.75 & 38.46 & 66.66 & 67.90 & 59.77   \\
& $\mathcal{L}_{adv}^{lm} +\mathcal{L}_{\text{adv}}^{dec}$ &  62.12 & 44.57 & 71.57 & 75.33 & 63.44 \\

& $\mathcal{L}_{adv}^{lm} +\mathcal{L}_{\text{adv}}^{dec} + \mathcal{L}_{\text{adv}}^{h}$ & 67.57 & 49.44 & 76.55 & 79.66 & 69.76 \\

& $\mathcal{L}_{adv}^{lm} +\mathcal{L}_{\text{adv}}^{dec} + \mathcal{L}_{\text{adv}}^{h} + \mathcal{L}_{adv}^{ctg}$ & {72.44} & {56.79} & {81.10} & {84.74} & {74.10}  \\

% & \textbf{LAMP} & \textbf{75.21}          & \textbf{57.43}          & \textbf{83.56}          & \textbf{87.23}          & \textbf{77.81} \\

\cmidrule{1-7}
% & Clean(No attack) & 45.62   & 58.88 & 39.55 & 54.80 &  40.90 \\
\multirow{6}{*}{\centering LLaVa-v1.6} & CPGC-UAP \cite{cpgc} & 59.76 & 48.74 & 67.55 & 49.09 & 63.35     \\
 & UAP-VLP \cite{uap_vlp}  & 60.79 & 48.66 & 68.66 & 50.10 & 64.44\\
 
 & Doubly-UAP &  51.43 & 29.64 & 52.34    & 42.53          & 55.31   \\
 & Jailbreak-MLLM & 36.15          & 21.33          & 45.43          & 38.43 & 40.13   \\
 
& $\mathcal{L}_{adv}^{lm}$  & 64.76 & 54.79 & 72.78 & 55.44 & 67.77   \\
& $\mathcal{L}_{adv}^{lm} +\mathcal{L}_{\text{adv}}^{dec}$ & 66.68 & 57.44 & 75.15 & 59.76 & 69.88 \\

& $\mathcal{L}_{adv}^{lm} +\mathcal{L}_{\text{adv}}^{dec} + \mathcal{L}_{\text{adv}}^{h}$ & 71.47 & 59.65 & 79.55 & 67.66 & 74.22  \\

& $\mathcal{L}_{adv}^{lm} +\mathcal{L}_{\text{adv}}^{dec} + \mathcal{L}_{\text{adv}}^{h} + \mathcal{L}_{adv}^{ctg}$  & {76.79} & {64.22} & {84.33} & {74.55} & {79.24}  \\ 

% & \textbf{LAMP} & \textbf{79.65}          & \textbf{67.51}          & \textbf{86.79}          & \textbf{77.55}          & \textbf{82.78} \\

\cmidrule{1-7}

% & Clean(No attack) & 39.17   & 58.72 & 31.17 & 45.90 &  42.15 \\
\multirow{6}{*}{\centering Qwen-VL-Chat} & CPGC-UAP \cite{cpgc} & 64.68 & 49.68 & 72.11 & 59.74 & 61.09   \\
 & UAP-VLP \cite{uap_vlp} & 66.68 & 52.11 & 74.22 & 63.29 & 65.55 \\

 & Doubly-UAP &  49.42 & 25.44 & 50.43    & 40.51          & 53.31   \\
 & Jailbreak-MLLM & 33.15          & 20.33          & 43.43          & 33.34 & 38.32   \\
 
& $\mathcal{L}_{adv}^{lm}$   & 69.66       & 56.57 & 76.66 & 67.44   & 68.22  \\
& $\mathcal{L}_{adv}^{lm} +\mathcal{L}_{\text{adv}}^{dec}$ & 72.11 & 59.57 & 79.79 & 70.59 & 71.44  \\

& $\mathcal{L}_{adv}^{lm} +\mathcal{L}_{\text{adv}}^{dec} + \mathcal{L}_{\text{adv}}^{h}$ & 77.55 & 62.55 & 81.08 & 72.13 & 75.68 \\

& $\mathcal{L}_{adv}^{lm} +\mathcal{L}_{\text{adv}}^{dec} + \mathcal{L}_{\text{adv}}^{h} + \mathcal{L}_{adv}^{ctg}$ & {81.35} & {64.63} & {84.68} & {74.66}  & {79.55}    \\ 

% & \textbf{LAMP} & \textbf{84.57}          & \textbf{67.15}          & \textbf{87.65}          & \textbf{76.12}          & \textbf{83.84} \\

\cmidrule{1-7}

% \multirow{6}{*}{\centering Qwen-2.5} & CPGC-UAP \cite{cpgc} & 63.84 & 48.83 & 71.14 & 58.41 & 60.94   \\
%  & UAP-VLP \cite{uap_vlp} & 65.45 & 50.15 & 72.24 & 60.23 & 64.59 \\

%  & Doubly-UAP &  48.41 & 23.42 & 48.43    & 38.13          & 50.31   \\
%  & Jailbreak-MLLM & 30.76  & 19.54          & 41.34          & 30.43 & 36.23   \\
 
% & \textbf{LAMP} & \textbf{82.71}          & \textbf{66.64}          & \textbf{87.65}          & \textbf{77.56} & \textbf{82.43} \\
% \bottomrule

\end{tabular}%
}
\caption{Attack success rate (\%) on the multi-image benchmarks compared with baselines and variants of our method.}
\label{tab:aux_table}
\end{table*}

\section{Qualitative analysis for position invariant attack}

\begin{figure*}[ht]
  \centering
  \begin{subfigure}[t]{0.45\textwidth}
    \centering
    \includegraphics[width=\linewidth]{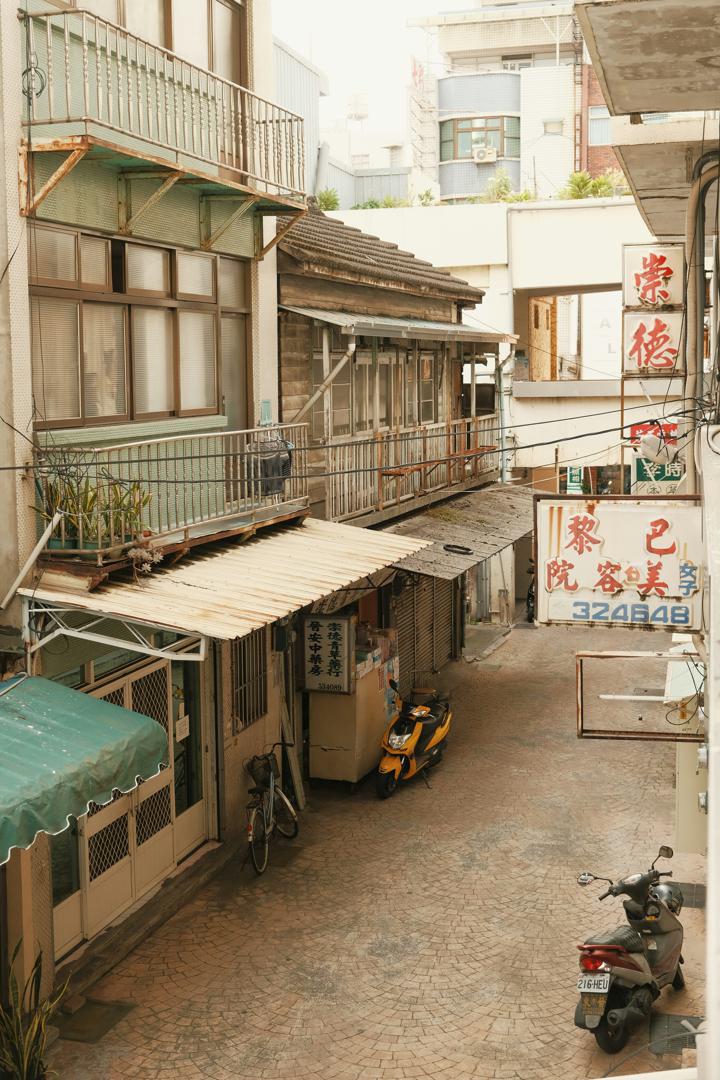}
  \end{subfigure}
  \begin{subfigure}[t]{0.45\textwidth}
    \centering
    \includegraphics[width=\linewidth]{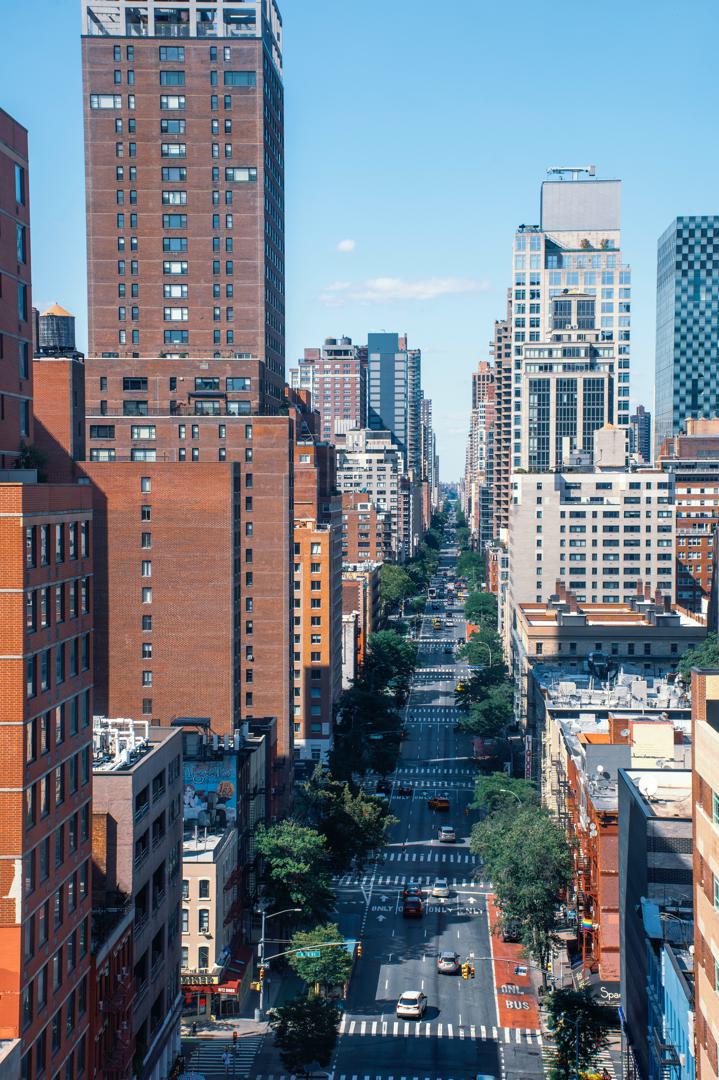}
  \end{subfigure}
   \caption*{\textbf{Original Input:} \\
    \textit{Prompt:} In (image 1: <Image><image></Image>) and (image 2: <Image><image></Image>), which image shows a more economically advanced place?\\
    \textit{Options:} (A) Image 1 is more developed; (B) Image 2 is more developed; (C) No, both images are equally developed.\\
    \textbf{Attacked Prediction:} A \quad \textbf{GT:} B}
    \caption{Qualitative analysis showing position sensitivity of MLLMs. This is the original ordering of images.}
    \label{fig:actual_position}
    
  % \hfill
\end{figure*}

\begin{figure*}[ht]
  \centering
  \begin{subfigure}[t]{0.45\textwidth}
    \centering
    \includegraphics[width=\linewidth]{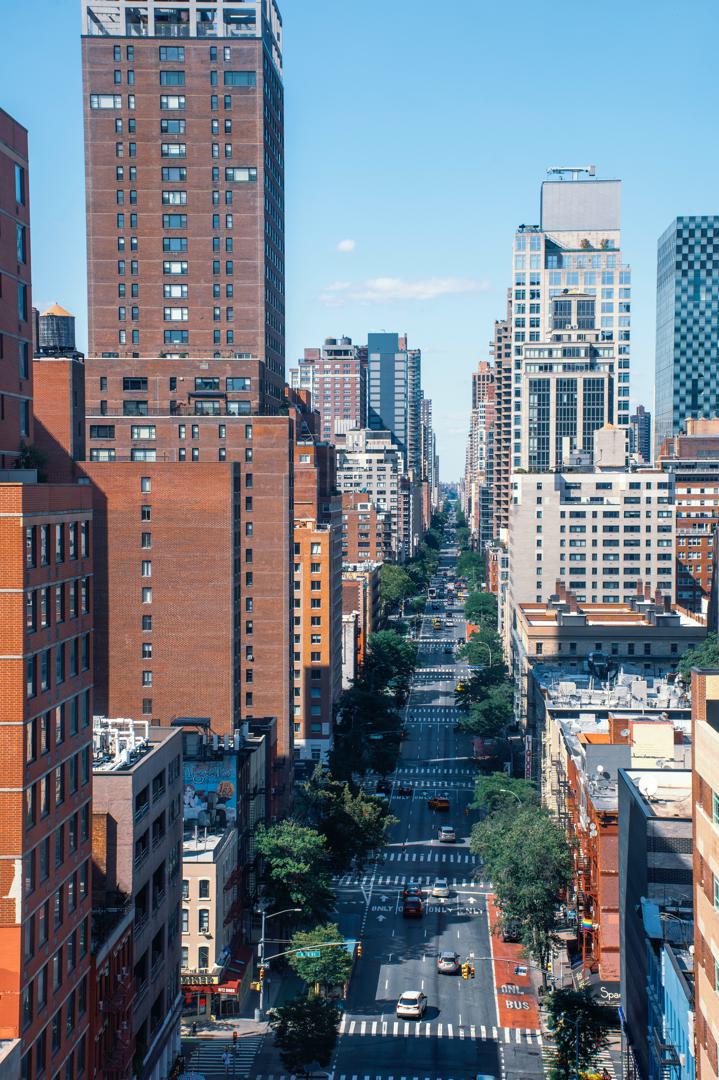}
  \end{subfigure}
  \begin{subfigure}[t]{0.45\textwidth}
    \centering
    \includegraphics[width=\linewidth]{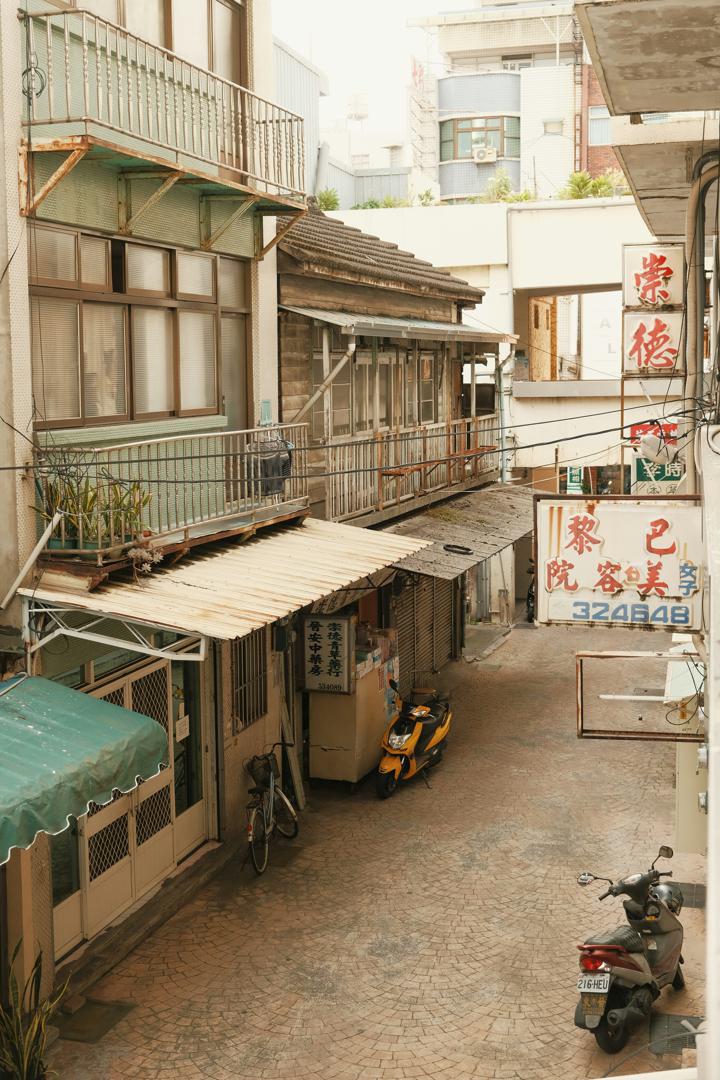}
  \end{subfigure}
   \caption*{\textbf{Flipped Input:} \\
    \textit{Prompt:} In (image 1: <Image><image></Image>) and (image 2: <Image><image></Image>), which image shows a more economically advanced place?\\
    \textit{Options:} (A) Image 1 is more developed; (B) Image 2 is more developed; (C) No, both images are equally developed.\\
    \textbf{Attacked Prediction:} B \quad \textbf{GT:} A}
  \caption{Qualitative analysis showing position sensitivity of MLLMs. The model’s prediction flips when the image order is reversed, despite the semantic content remaining unchanged.}
  \label{fig:position_flip}
\end{figure*}

We are showing the qualitative analysis of the position invariant attack, that is achieved by imposing index-attention suppression constraint. When the learned perturbation is added to the images, the model incorrectly picks the answer in Fig. \ref{fig:actual_position}. When the position is changed, the output is changed in Fig. \ref{fig:position_flip}, and still picks the incorrect answer.

\section{Effectiveness LAMP across vision encoders}

We do not impose any constraints based on the vision encoders; our method is fundamentally designed around the LLM backbones. Most MLLMs leverage popular architectures similar to LLaMA, where decoder layers incorporate multi-head self-attention mechanisms. As a result, our approach is applicable across a wide range of open-source foundation models, including LLaMA, Mantis, VILA, Qwen, and LLaVA.

Our adversarial language modeling loss constrains the LM head, while our adversarial hidden states and adversarial attention (using the Pompeiu-Hausdorff distance) impose constraints on LLM decoder layers. Finally, the contagious loss targets the self-attention mechanism within the LLM decoders. Because most open-source MLLMs share a similar LLM backbone—regardless of the specific vision encoder architecture—our approach is both reproducible and transferable across various MLLMs.

\section{Additional defense result}

We compare LAMP with  \cite{x_transfer} that is a super transferable attack. LAMP shows better performance for both defense method across attack method and datasets in Tab. \ref{tab:defense_method}.

\begin{table}[]
    \centering
    \begin{tabular}{c|c|c}
    \toprule
         Defense & Method &  ASR \\
         \midrule
         \cite{random_noise}    & LAMP &  70.23\%\\
        \cite{blacklight}  & LAMP  & 69.21\%\\
         \bottomrule
         \cite{random_noise}    &  \cite{x_transfer}   &  56.33\%\\
         
         \cite{blacklight}    & \cite{x_transfer}  & 20.21\%\\

        \bottomrule
    \end{tabular}
    \caption{ASR against blackbox defense strategies on Mantis Eval dataset and Mantis-CLIP model}
    \label{tab:defense_method}
\end{table}

\section{Experiments with selection free VQA, Image captioning}

In the main paper, we present the multi-image tasks. Now, we present the selection free VQAv2 tasks and image captioning tasks following \cite{pandoras_box}.  we present results for selection-free VQAv2 and image captioning tasks (MS-COCO). Specifically, we measure and report the semantic similarity scores between the MLLMs' outputs and the attacker's chosen label, ``Unknown''.

\begin{table*}[]
    \centering
    \begin{tabular}{c|c|c|c}
    \toprule
         Method  & Model & Image captioning & VQA  \\
         \toprule
         % MF Attack & LLaVA  0.616 & 0. 0.667 \\
         % Pandora's box & LLaVA  &   0.812 & 0.828 \\
        X-transfer \cite{x_transfer}  & Mantis-CLIP & 23.2 & 29.5 \\
         LAMP & Mantis-CLIP & \textbf{ 72.6} & \textbf{73.4} \\
         \midrule
         X-transfer & LLaVA & 29.6 & 31.5 \\
         LAMP & LLaVA & \textbf{ 73.1} & \textbf{74.5} \\
         \midrule
         %  MF Attack & MiniGPT4  0.614 & 0.671 \\
         % Pandora's box & MiniGPT4  &   0.826 & 0.851  \\
         X-transfer & MiniGPT & 28.9 & 30.1 \\
         LAMP & MiniGPT4 & \textbf{78.3} & \textbf{80.7} \\
         \bottomrule
    \end{tabular}
    \caption{MS COCO and Ok-VQA attack performance with X-transfer, ASR (\%) results in image captioning and VQA across various large VLMs and datasets. For image captioning,
CIDEr is used as the evaluation metric, while VQA accuracy is employed for the VQA task following X-transfer}
    \label{tab:my_label}
\end{table*}

\section{Complexity analysis}

For synthesizing the most effective attack, we optimize the perturbation(s) using 17,000 training samples from the Mantis Instruct dataset. The primary computational and memory overheads arise from querying the victim MLLMs during the perturbation learning process. The perturbations are overlaid on a subset of images during each iterative update, constrained by a fixed perturbation budget. 

It is important to note that our attack is a \textit{universal} multi-image attack against multi-image MLLMs. This means that the same universal adversarial perturbations can be applied to any image from any task to achieve a successful attack, in contrast to existing approaches that require generating a separate perturbation for each input sample. Tab. \ref{tab:complex}

\begin{table}[]
    \centering
    \begin{tabular}{c|c}
    \toprule
         Process & Avg. GPU hours \\
         \midrule
         Gradient update for all loss function  &  48h\\
         \bottomrule
         
    \end{tabular}
    \caption{Complexity in training}
    \label{tab:complex}
\end{table}

We learned a set of UAP(s) using a source MLLM. During inference, the perturbations are overlaid across test samples to generate adversarial examples. Note that, the attacker has no control over test samples to learn the UAP(s). Adversarial samples using UAP-unimodal are also generated using a similar approach. We include the test-time performance for generating a single adversarial example. Jailbreak-MLLM focuses on a sample-level attack. In other words, it optimizes each image individually, and need to have access to test images. Jailbreak-MLLM uses the term "universal" because each perturbed image elicits diverse harmful-yet-helpful outputs from the attacked VLM(s). However, it does not learn a truly "universal" perturbation as defined in previous studies. Tab. \ref{tab:time}

\begin{table}[]
    \centering
    \begin{tabular}{c|c}
    \toprule
         Method & time (sec)\\
         \midrule
         UAP-Unimodal &  0.5 \\
         Jailbreak-MLLM & 600 \\
         LAMP & 0.5\\
         \bottomrule
    \end{tabular}
    \caption{Comarison for test time performance}
    \label{tab:time}
\end{table}

\section{Sensitivity of hyperparameters}
The weights of all loss functions are set to 1, except for the contagious and position-invariant losses, which are set slightly higher at 1.2. This is done to encourage the model to depend slightly more on these two loss terms. However, we observe that if all the loss function weights are uniformly set to 1, the results do not change significantly. We report the Attack Success Rate (ASR) on the Mantis CLIP and Mantis Eval benchmarks for different combinations of loss weights. These results show that setting all the loss weights to 1 does not significantly impact performance. Our results suggest that the five loss functions are not highly sensitive to weight tuning and perform well without extensive hyperparameter optimization, making them easily adoptable in theory and practice.

\section{Imperceptibility quantification.} In practice, we empirically showed that the learned perturbations are highly transferable across different model architectures and tasks. In other words, the perturbations need to be synthesized across different architectures every time to attain an imperceptible attack success rate. Tab. \ref{tab:hyper}
\begin{table}[]
    \centering
    \begin{tabular}{c|c}
    \toprule
           Hyperparameters & ASR  \\
           \midrule
          All $\lambda = 1$   & 72.9 \\
          $\lambda_1, \lambda_2, \lambda_3 =1 $, $\lambda_4, \lambda_5 = 1.2$  & 73.43 \\
          \bottomrule
    \end{tabular}
    \caption{Hyperparametr sensitivity}
    \label{tab:hyper}
\end{table}

\section{Interaction between losses.}
We introduce a contamination index $$
CI = \frac{1}{|C||N|} \sum_{i \in C} \sum_{j \in N}
\mathrm{Attn}_{i \rightarrow j} \cdot \mathrm{cosdist}(h_i, h_j),$$

which measures the attention-weighted divergence from clean tokens to perturbed tokens; higher values indicate a stronger adversarial influence.

On Mantis-Eval, without $\mathcal{L}_{\text{ctg}}$, we obtain $CI = 0.14$ and an attack success rate (ASR) of $61.1$. With $\mathcal{L}_{\text{ctg}}$, the contamination index increases to $CI = 0.31$, and ASR improves to $73.4$.

We compute the cosine similarity between gradients induced by different loss terms. The gradient similarity between $\mathcal{L}_{\text{ctg}}$ and the remaining losses is $0.0032$, while that of $\mathcal{L}_{\text{ias}}$ against others is $0.0021$, indicating minimal interference among objectives.

Beyond attack success rate (ASR), we additionally report CIDEr and accuracy (for captioning and VQA, respectively), as well as the change in CLIPScore, where $\Delta$CLIPScore = $-0.33$.

% \bibliography{aaai2026}